\documentclass{article}

     \PassOptionsToPackage{numbers, compress}{natbib}

 \usepackage[preprint]{neurips_2026}

\usepackage[utf8]{inputenc} 
\usepackage[T1]{fontenc}    
\usepackage{hyperref}       
\usepackage{url}            
\usepackage{booktabs}       
\usepackage{amsfonts}       
\usepackage{nicefrac}       
\usepackage{microtype}      
\usepackage{xcolor}         
\usepackage{outlines, amsmath, minted, algorithm, algpseudocode, graphicx, wrapfig, multirow, caption, enumitem, wrapfig}
\usepackage{mdframed}
\usepackage[nodisplayskipstretch]{setspace}

\usepackage{pifont}
\newcommand{\cmark}{\ding{51}}
\newcommand{\xmark}{\ding{55}}

\usepackage[textsize=scriptsize]{todonotes}
\usepackage{inconsolata}

\newcommand{\method}{{\sc Vesta}}
\newcommand{\methodfull}{\textbf{V}isual \textbf{E}xploration with \textbf{S}tatistical
\textbf{T}ool \textbf{A}gents}
\newcommand{\data}{{\sc Dawn}}
\newcommand{\datafull}{\textbf{D}ataset for \textbf{A}utomated \textbf{W}orkflows and \textbf{N}umerical Modeling}

\usepackage{xcolor}
\usepackage{tikz}
\usepackage{fontawesome5}

\definecolor{easygreen}{RGB}{60, 140, 80}
\definecolor{easyfill}{RGB}{220, 245, 225}
\definecolor{medyellow}{RGB}{180, 140, 20}
\definecolor{medfill}{RGB}{250, 240, 200}
\definecolor{hardred}{RGB}{180, 50, 50}
\definecolor{hardfill}{RGB}{250, 220, 220}
\definecolor{astroblue}{RGB}{40, 110, 170}
\definecolor{astrofill}{RGB}{230, 240, 255}

\newcommand{\splitlabel}[3]{%
  \tikz[baseline=-0.6ex]{%
    \node[
      rounded corners=3pt,
      fill=#2,
      draw=#1,
      line width=0.8pt,
      inner xsep=4pt,
      inner ysep=2pt,
      font=\bfseries\small
    ] {#3};%
  }%
}

\usepackage[skins,breakable]{tcolorbox}
\usepackage{tikz}
\usetikzlibrary{tikzmark,decorations.pathreplacing,calc}

\definecolor{light-purple}{RGB}{151,156,171}
\definecolor{blue-color}{RGB}{40,166,189}
\definecolor{pink-color}{RGB}{237,46,104} 
\definecolor{dark-grey-color}{RGB}{79,91,102}
\definecolor{darkbyzantium}{rgb}{0.36, 0.22, 0.33}
\definecolor{bluebell}{rgb}{0.64, 0.64, 0.82}
\definecolor{airforceblue}{rgb}{0.36, 0.54, 0.66}
\definecolor{response}{RGB}{245,198,165}

\newtcolorbox[list inside=prompt,auto counter,number within=section]{prompt}[1][]{
    colbacktitle=airforceblue,
    colframe=airforceblue,
    fontupper=\footnotesize,
    boxsep=5pt,
    left=0pt,
    right=0pt,
    top=0pt,
    bottom=0pt,
    boxrule=1pt,
    enhanced, 
    breakable,
    skin first=enhanced,
    skin middle=enhanced,
    skin last=enhanced,
    #1,
}

\newcommand{\Easy}{\splitlabel{easygreen}{easyfill}{Easy}}

\newcommand{\Hard}{\splitlabel{hardred}{hardfill}{Hard}}
\newcommand{\Astro}{\splitlabel{astroblue}{astrofill}{Astro}}

\title{\method{}: Visual Exploration with Statistical Tool Agents}

\author{%
  William Rudman$^1$\thanks{Equal contribution.},{ } Abhishek Divekar$^{1,2*}$,{} Kanishk Jain$^{1*}$,{} Sebastian Joseph$^1$, \\
  \textbf{Stella S. R. Offner$^1$, Matthew Lease$^1$, Kyle Mahowald$^1$, Greg Durrett$^{3}$\thanks{Equal advising.} , Junyi Jessy Li$^{1\dagger}$} \\[0.5em]
  $^1$The University of Texas at Austin \quad $^2$Amazon \quad $^3$New York University \\[0.3em]
  \small Correspondence: \texttt{william.rudman@utexas.edu} \, | \,
  \faGithub\ \small\url{https://github.com/wrudman/VESTA}
}

\begin{document}

\maketitle

\begin{abstract}
Fitting quantitative models to data is a central step in scientific workflows, yet it remains one of the least automated. Recent agent-based systems leverage language and vision-language models (VLMs) to propose statistical models, but these systems struggle on challenging modeling tasks. To address these limitations, we introduce \method{} (\methodfull{}), a framework that equips VLMs with a dynamically growing toolkit to guide model refinement through data transformations, hypothesis-driven visualizations, and statistical tests. Unlike prior systems that rely on iterative critique alone, \method{} explores data before and during refinement by selecting or creating reusable diagnostic tools. We evaluate \method{} using three toolkit configurations: no tools, static expert-written tools, and dynamic model-written tools. 
To support this evaluation, we introduce \data{} (\datafull{}), a benchmark targeting distribution fitting and time series modeling with varying difficulty tiers that culminates in real-world astronomy tasks including modeling initial mass functions and gravitational-wave chirp signals. \method{}'s dynamic tool creation outperforms prior agentic pipelines for statistical analysis and visual tool creation, with the largest gains on complex and domain-specific tasks. Against strong agentic coding harnesses (Claude Code, Codex, and OpenCode), \method{} is better on all time series tasks and on complex distribution fitting tasks, though the harnesses achieve better fits on simpler distributions. Further, dynamically generated tools are substantially more sophisticated than those produced by existing visual tool-creation systems, covering more diagnostic categories per function and strongly preferring visual outputs that the VESTA VLM critic can reason over directly.
\end{abstract}

\section{Introduction} 
Large Language Model (LLM) and Vision-Language Model (VLM) agents have demonstrated remarkable promise in scientific discovery, with systems capable of forming hypotheses, designing experiments, and synthesizing results. However, failures at different steps impede these models' abilities to truly accelerate science. This work focuses on the task of \textit{fitting quantitative models to data}. This step is complex, requiring priors about the domain in question and careful statistical analysis, while being less studied than other quantitative analysis tasks like ML engineering and data science \citep{mle_bench, dab_step}.
Addressing this gap is particularly important in scientific disciplines like astronomy where models struggle to effectively use domain conventions \cite{joseph2025astrovisbench}. 

We introduce \method{}\footnote{VESTA is the second most massive body in the main asteroid belt \cite{NASA4Vesta}.} (\methodfull{}), a framework that proposes, refines, and evaluates statistical models by analyzing \textit{visual representations} of data. 
\method{} operates over models defined in PyMC \cite{pymc}, a Bayesian probabilistic programming framework that enables model definition and fitting to data using an efficient Markov Chain Monte Carlo (MCMC) method. Model-building with PyMC code is compositional, allowing for complex distributions to be constructed by combining simpler components. 
\method{} instantiates a loop of proposing models, selecting the best one, then iterating, broadly similar to existing autoresearch workflows \citep{ada_evolve, execution_grounded_research}. However, unlike prior work, \method{} utilizes multimodal feedback by augmenting this basic loop 
with a newer variety of \textbf{visual tools}. Visual tools allow \method{} to inspect model fits using a VLM, giving it leverage in a modality distinct from past work on tool use. 

\begin{figure}[t !]
    \centering
    \includegraphics[width=\linewidth]{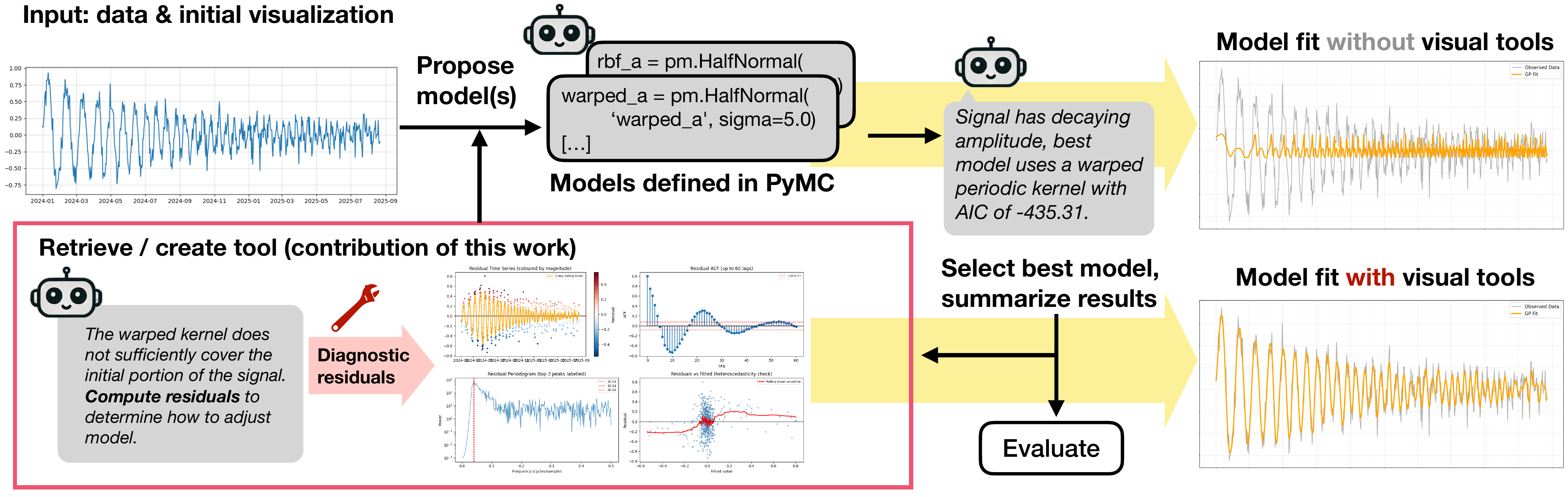}
    \caption{Overview of \method{}. By effectively using and creating tools, \method{} produces a probabilistic PyMC program that models the input data. 
}
    \label{fig:intro-fit}
\end{figure}

Figure~\ref{fig:intro-fit} shows how a model can use a tool to compute residuals, produce a visualization, and then inspect that visualization to identify areas where the predicted model does not fit the data well and propose a new statistical model that better fits the data. For the diversity of statistical model fitting tasks we explore here, it is infeasible to produce all necessary tools beforehand;  \emph{dynamic} tool instantiation is required to cover the full breadth of cases. 

To support advancement on challenging model fitting problems where current approaches fail, we introduce a new benchmark 
we call \data{} (\datafull{}).\footnote{DAWN was NASA’s first truly interplanetary spaceship that explored VESTA \cite{NASADawnMission}.} \data{} problems cover two domains central to both general data science and scientific research: fitting probability distributions to discrete data and constructing models for time-series analysis. 
In contrast to existing benchmarks which often contain a small number of distributions with a limited number of points \cite{boxing_gym, lloyd2014automaticconstructionnaturallanguagedescription}, 
\data{} consists of synthetic problems spanning two levels of difficulty (\Easy{} and \Hard{}), as well as 
two fundamental problems from astronomy (\Astro{}): fitting stellar initial mass functions (IMFs), and detecting and characterizing chirp signals embedded in gravitational-wave data. 

Using \data{}, we benchmark \method{} against prior frameworks specialized for Bayesian modeling and dynamic VLM-based tool-generation systems. While existing approaches and coding harnesses perform competitively on our \Easy{} splits, performance degrades substantially as data complexity increases and tasks shift toward domain-specific scientific phenomena. This degradation reveals a systematic gap in current agents' capacity for iterative, evidence-driven model refinement. 
\method{} recovers expert-created tools at a higher rate than prior work, creating sophisticated tools that compose several expert tools, testing multiple hypotheses simultaneously. While dynamically generated visual tools do not outperform expert-written tools, likely due to current agents' limited ability to process complex diagrams. \method{} shows great promise at producing on-demand statistical tooling and we expect effectiveness will improve alongside rapidly evolving VLM capabilities.

\section{Related Work}
\paragraph{Automated Model Fitting} Past work evaluated the capabilities of LLMs and VLMs to automate statistical analysis \citep{llm_statisticians, llm_stat_overview}, time-series modeling \citep{time_series_transformer, transformer_ts_survey, llm_zero_shot_ts, time_vlm, vlm_ts_survey} and Bayesian model proposals \citep{llm_processes, critical}. \citet{llm_stat_overview} and \citet{llm_statisticians} survey how LLM-based agents can automate statistical workflows from data cleaning and model selection. However, \citet{are_llms_good_statisticians} find that even state-of-the-art LLMs achieve only modest accuracy on benchmarks for selecting appropriate statistical methods, revealing significant limitations in LLMs' ability to reason over complex statistical tasks. While much of the focus in the literature has used LLMs / VLMs to predict a continuation of time series or detect anomalies, our work seeks to represent the \textit{underlying model} that generates the observed data using a probabilistic Bayesian model (PyMC). Previous work in generating PyMC code to model data has been successful in scientific settings, but this work often relies on human-written kernels for Gaussian processes, limiting the use cases of such methods \cite{amd_non_linear_dynamics, probabilistic_grammars, kernel_search}. Recent work addresses these key limitation by using LLMs to automatically perform ``Box's Loop'' of iteratively proposing new  models and critiquing existing ones written in PyMC code until an adequate fit is achieved \cite{box_lm}. Despite some promising results using LLMs to propose probabilistic models, \citet{boxing_gym} find LLMs struggle on more challenging datasets sampled from real-world domains and demonstrate that the iterative refinement of PyMC models does not consistently provide better predictions over one-shot LLMs. Further, while existing benchmarks only cover a narrow range of distributions, \data{} combines a variety of complex mixture distributions inspired by real world tasks. Our \method{} automates Box's propose-critique-refine loop using VLMs, enabling tool use and creation for data analysis, while handling significantly more complex distributions.

\paragraph{Multimodal Tool Use \& Creation} Several works have demonstrated that VLMs are proficient tool-users and that the successful utilization of tools can improve fine-grained reasoning  \citep{visual_sketchpad, image_of_thought, visual_chagpt, vis_mem}. Incorporating VLM tool-use in agentic systems has produced even larger performance gains \citep{skywork, deep_eyes, act_wisely, codev}. In particular, \citet{vipact} propose a multi-agent VLM framework in which an orchestrator agent coordinates specialized sub-agents and vision expert tools to improve fine-grained visual perception. In PyVision \cite{pyvision}, VLMs are used in an agentic flow where they dynamically create and execute tools until an answer is reached. Previous work has limited VLM tool use (i.e., executing or generating code over visual inputs) to image enhancements, annotations, cropping, and the calculation of basic statistics \cite{visual_programming, viper_gpt, visual_perception_w_tools}. While some work has applied multimodal tool creation to model fitting \cite{multimodal_amd}, such tools are limited to simple statistics and basic plotting, lack a dedicated tool-creation step, and are discarded after each use rather than persisted for reuse across iterations. In contrast, \method{} uses an LLM tool agent to dynamically create visualizations that test statistical hypotheses, which are then passed to a VLM critic as evidence for refining them.

\paragraph{Autonomous Tool Expansion} While both LLMs and VLMs have proven to use tools effectively, finding the optimal set of tools for a problem often requires manually curating a toolkit. Initial work demonstrates that LLMs are capable of coding effective and reliable tools \cite{creator, latm}, although these tools are typically created separately from model inference \cite{evoskill, tookmaker}. Systems such as Voyager \cite{voyager} extend this static paradigm by allowing for tool libraries to adapt to task needs and evolve across examples. Many of the systems for autonomous tool expansion tend to focus on software engineering tasks \cite{evoclaw, sea-agent, open-sage} or narrowly scoped API invocation benchmarks \cite{toolace, rest-gpt}, with less focus on problems that require judgment-driven statistical analysis. While EvoSkill \cite{evoskill} and TTE \cite{tte} create some tools related to calculating statistics, they tend to contain more simplistic functions such as calculating averages, areas or retrieving information from tables.   

\section{\method{}: Visual Exploration  with Statistical Tools}

We assume as input a dataset $D$ generated by a model $M(\theta)$. $M$ is a data generating process (e.g., a directed graphical model) with parameters $\theta$. In this work, we consider models that can be specified in PyMC code. Given a proposed model $\tilde{M}(\tilde{\theta})$, we can evaluate its fit to $D$ with metrics denoted as $R(D,\tilde{M}(\tilde{\theta}))$. Central to our approach are \textit{visual tools}, $\varepsilon(D, M, \theta)$, implemented as executable Python functions that accept the data $D$ and a current model $(M, \theta)$ and return visual or statistical diagnostics used to inform subsequent modeling decisions.

\subsection{\method{}}
Algorithm~\ref{alg:vea} illustrates the workflow for \method{}. We start with an initial prediction by estimating a probabilistic model from an initial plot of the data, and then refine this prediction over $N$ steps. At each step, we propose $p$ alternative candidate models, then select the model with the best metric, $R$. Our approach proceeds in four steps per iteration: propose, 
generate tools, critique model fit, and summarize steps of the run.

\textbf{Propose.} On the first iteration, \method{} takes an initial 
visualization of the data, produces a brief visual description  and is instructed to create $p$ diverse model proposals. To ensure faithfulness
to the proposed model parameters, PyMC code is generated in a dedicated step that 
translates parameter specifications into executable PyMC programs. Without this step, we find models frequently deviate from their specified 
parameterizations.

\begin{wrapfigure}{r}{0.59\textwidth}
\begin{minipage}{0.58\textwidth}
\begin{algorithm}[H]
\caption{VESTA}
\label{alg:vea}
\begin{algorithmic}[1]
\Require Data $D$, iterations $N$, proposals $p$, metric $R$, registry $\mathcal{E}$ (initial state: \texttt{generate\_new\_tool} only)
\Ensure $M_{\text{best}}, \theta_{\text{best}}$
\State $M_{\text{best}}, \theta_{\text{best}} \gets \textsc{Propose}(\text{Plot}(D))$ \Comment{Plot data; fit $j$ models and select best model under metric, $R$}
\State $s_0 \gets \textsc{Summarize}(M_{\text{best}}, \theta_{\text{best}})$
\For{$i = 1, \ldots, N$}
    \State $\varepsilon^* \gets \textsc{ToolManager}(s_{i-1}, D, \mathcal{E})$ \Comment{Select/create tool; update $\mathcal{E}$}
    \State $T_i \gets \varepsilon^*(D, M_{i-1}, \theta_{i-1})$ \Comment{Execute tool}
    \State $M_i, \theta_i \gets \textsc{Critique}(T_i, s_{0:i-1}, D)$ \Comment{Propose $p$ models; return best ranked by $R$}
    \State $M_{\text{best}}, \theta_{\text{best}} \gets \arg\min_{\theta \in \{\theta_i, \theta_{\text{best}}\}} R(\mathcal{M}(\theta))$
    \State $s_i \gets \textsc{Summarize}(M_i, \theta_i)$
\EndFor
\State \Return $M_{\text{best}}, \theta_{\text{best}}$
\end{algorithmic}
\end{algorithm}
\end{minipage}
\end{wrapfigure}

\textbf{Generate Tools.} The central contribution of \method{} is its ability to explore data through the composition and creation of tools. At each iteration, \method{} receives a summary of the previous runs then decides to either select an existing tool from a growing registry $\mathcal{E}$, or dynamically codes a new tool in Python when no existing tool is adequate for the desired analysis. 
Newly created tools are saved to the registry, making them reusable for the current run. At each run, the toolkit is newly initialized. Section~\ref{sec:tools} presents a case study of generated tools (Figure~\ref{fig:tool_ex}).

\textbf{Critique.} The output of the selected or newly created tool, along with the history of the best models and descriptions from previous runs is passed to a VLM critic that refines the current PyMC model by proposing $p$ revised models. The revised model that optimizes $R$ is passed to the next iteration. In our prompt, we instruct the VLM critic to terminate the iterative refinement 
loop early if it judges the current model fit to be satisfactory. The critic signals this with a \texttt{COMPLETE} token.

\textbf{Summarize.} At the end of each iteration, the visual description, 
chosen model and parameters, the metric $R$, and tool output are compressed into a 
structured summary. This summary is injected into the next iteration in place 
of raw intermediate outputs, enabling the system to reason over the full 
refinement trajectory without incurring prohibitive context growth.

\subsection{Visual Tools in \method{}}\label{sec:vestatools}

\method{} can work out-of-the-box without tools.  
However, we found tools to be critical in solving challenging model fitting tasks. 
In particular, we allow \method{} to propose new tools tailored to its model predictions. At each tool generation step, \method{} receives a summary of the previous iteration, including the VLM's critique of the model fit, the best PyMC model and its score $R(D,M_{\text{best}}(\theta))$, and a history of previously executed tools. The agent also has access to a registry of available tools.
We initialize the toolkit with a single function that gives the option for creating a new tool. New tools emerge through the process of examining data, and allow models to verify and update an existing hypothesis (e.g., a proposed statistical distribution).
We save successfully created tools. If a generated tool fails, we allow up to two retries, passing the failure trace into the retry prompt. After two retries if a tool fails to execute due to syntax or import errors, the system falls back to the ``no toolkit'' mode. 

In practice, \method{} generates complex tools that test distinct hypotheses through visualization, and the Critique stage uses their outputs (case study in Section~\ref{sec:tools}, Figure~\ref{fig:tool_ex}). \method{} can also use externally provided tools. Section~\ref{sec:impl} describes our \textbf{expert-written} toolkit, which we evaluate in Section~\ref{sec:results}.

\section{The \data{} Benchmark}
Distribution fitting and time series modeling are two key data science modeling tasks 
that appear consistently across scientific disciplines. We select these domains 
because they allow us to benchmark AI agents across easily recognizable models, 
complex mixtures of common forms, and challenging real-world tasks where 
domain-specific knowledge, such as established conventions and governing 
equations, can reduce the complexity of modeling real-world phenomena. \data{} problems are synthetically generated, allowing us to robustly measure model fit.

\begin{wrapfigure}{r}{0.53\textwidth}
  \centering
  \includegraphics[width=0.51\textwidth]{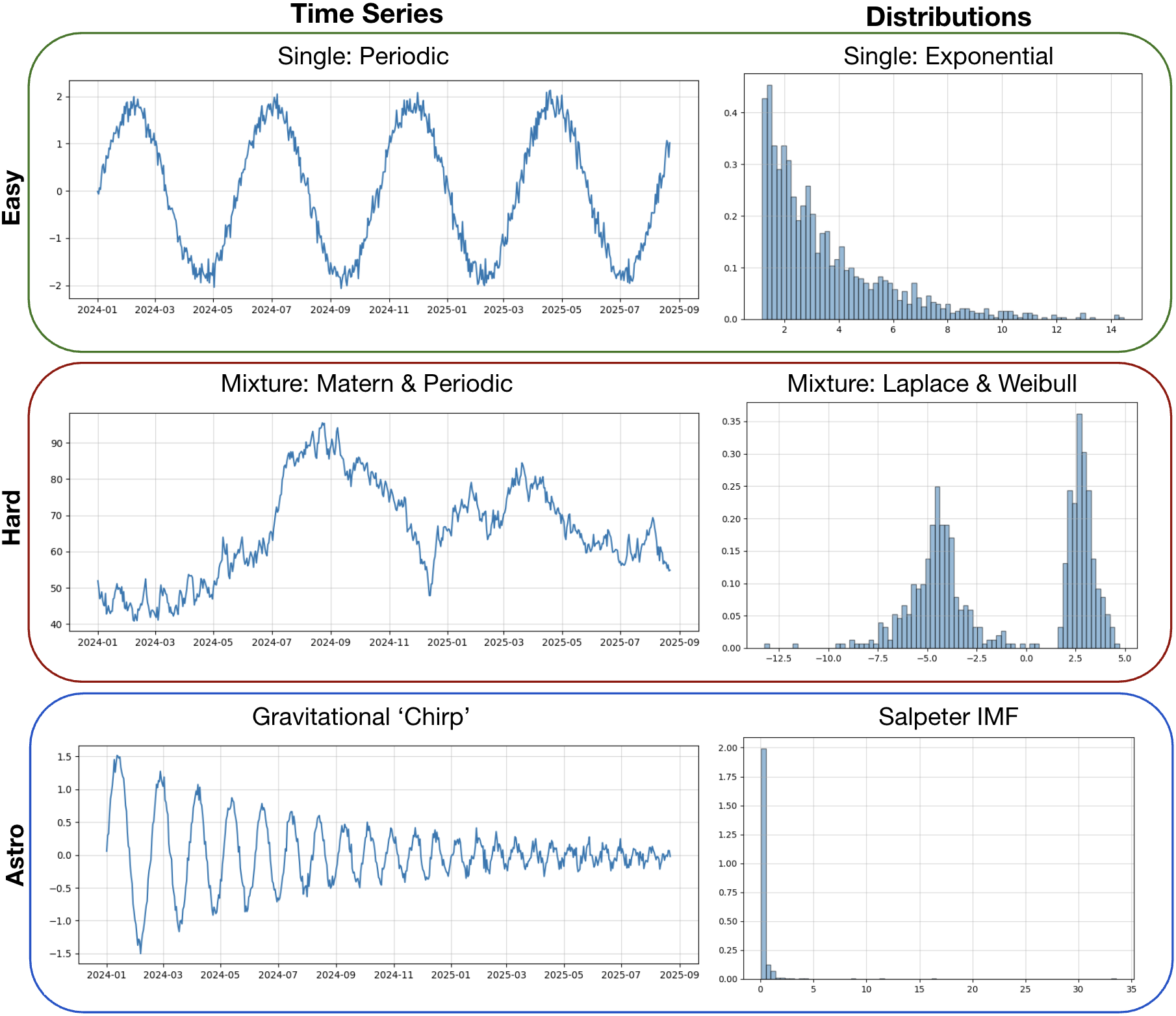}
  \caption{Sample inputs from both domains and all dataset splits in \data{}. \protect\Easy{} splits contain easily recognizable forms. \protect\Hard{} tasks contain mixtures of distinct forms, and \protect\Astro{} tasks reflect real-world astronomy challenges that require additional analysis beyond simple visualization to solve.}
  \label{fig:data_ex}
  \vspace{-5pt}
\end{wrapfigure}

\subsection{Distribution Fitting Tasks}
\vspace{-0.3em}
The \protect\Easy{} tasks in distribution fitting consist of identifying the family and associated parameters for a unimodal distribution. The distribution families we consider in this paper are: {\small{\tt gaussian}}, {\small{\tt lognormal}}, {\small{\tt student-t}}, {\small{\tt exponential}}, {\small{\tt uniform}}, {\small{\tt weibull}}, {\small{\tt laplace}}, {\small{\tt cauchy}} and {\small{\tt pareto}}. Both the priors and exact number of points for each distribution are randomly sampled. The \protect\Hard{} task composes exactly two of these distributions. Mixtures present in the \protect\Hard{} split model phenomenon present in other real-world domains such as the gaussian-laplace mixtures common in speech signal processing \cite{speech_signal} and lognormal-exponential mixtures prevalent in survival-analysis \cite{maller2024mixture}.

For our \protect\Astro{} split, we use initial mass functions (IMF) as the distribution family, which describe the distribution of stellar masses at birth and are often modeled with canonical forms \cite{imf_survey}. Variations in its form offer a window into past star-forming environments. Because the IMF is derived empirically, accurate statistical modeling is critical to scientific inference. Accurate modeling of the IMF allows astronomers to infer the conditions of past star formation and predict its downstream effects on galactic evolution and heavy element distribution~\citep{annurev-astro-082708-101642}.

\begin{figure}
    \centering
    \includegraphics[width=0.95\linewidth]{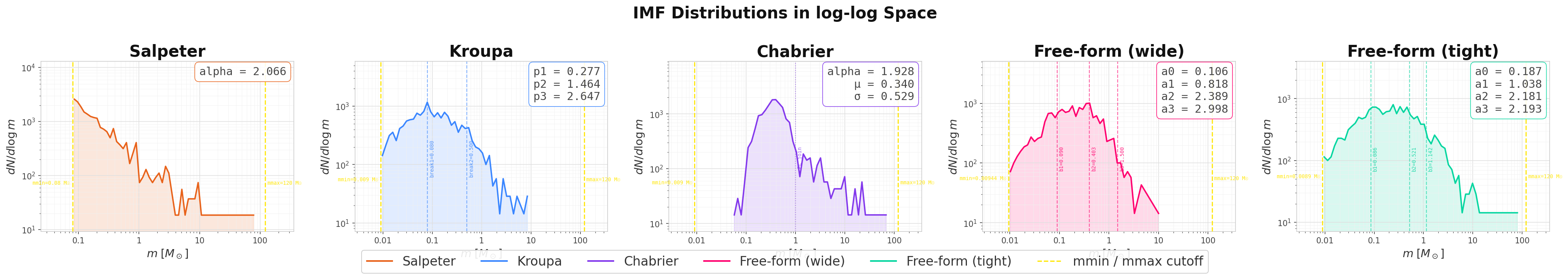}
    \caption{\data{}'s \protect\Astro{} distribution fitting tasks. Example of Initial Mass Functions projected into log-log space. Distributions become visually distinct only when projected into log-log space.}
    \label{fig:imf_ex}
    \vspace{-1em}
\end{figure}

We use five functional forms (visualized in Figure~\ref{fig:imf_ex}):
\textbf{(1)} The simplest form, the Salpeter IMF \cite{salpeter}, is a single power law.
\textbf{(2)} The Kroupa IMF \cite{Kroupa_2001} is a piecewise power law with sharp breaks at specific characteristic masses.
\textbf{(3)} The Chabrier IMF \cite{Chabrier_2003} is distinguished by a smooth log-normal component at low masses transitioning to a power law at higher masses. We additionally include two freeform variants that do not follow canonical forms, making them more challenging to model: 
\textbf{(4)} one with breaks constrained near canonical masses (freeform tight) and 
\textbf{(5)} one with unconstrained breaks and slopes that can produce more exotic stellar IMFs (freeform wide).

\paragraph{Evaluation Metric.}
\vspace{-0.3em}
We use Jensen-Shannon Divergence to measure similarity between our proposed model's distribution and the reference model's distribution. This metric is symmetric and bounded in the interval $[0, 1]$, with a score of 0 indicating identical distributions. Additionally, in Appendix~\ref{app:alterative_evals} we analyze the impact of alternative evaluation metrics and use a ``slope parameter recovery'' metric specifically designed for IMF fitting.  

\subsection{Time Series Fitting Tasks}
\vspace{-0.3em}
The \protect\Easy{}  task in time series fitting consists of modeling sequences characterized by simple linear trends (increasing, decreasing, or flat) and standard seasonal components, such as basic periodic sine waves, coupled with minimal noise. The \protect\Hard{} task introduces more complex, non-linear dynamics and non-standard periodicities. This split includes time series exhibiting random walks (ARIMA process), and S-curve (sigmoidal) functions. Further, \protect\Hard{} contains time series from other domains, such as electrocardiogram (ECG) signals of heartbeat patterns and time series exhibiting seasonal periodicity with increasing averages common in sales forecasting \cite{ensafi2022timeseries}. Capturing the behavior of these series requires moving beyond standard linear or single-period functions to employ multi-periodic modeling and Gaussian process kernels. This split tests the model's ability to handle richer structures and moderate levels of noise and non-linearity. 

For our \protect\Astro{} split, we model time series inspired by gravitational wave chirps. ``Chirps'' are an astronomical phenomenon produced by merging binary star systems that introduces severe non-stationarity into the signal. The standard chirp variant features a continuously changing frequency over time, where initial high-wavelength, low-frequency waves progressively compress into higher frequencies as the celestial bodies converge. We additionally include a highly complex variant where this dynamic, shifting frequency is coupled with an amplitude that decays over time.

\paragraph{Evaluation Metric.}
\vspace{-0.3em}
Similar to prior works \cite{box_lm}, we use the Expected Log Pointwise Predictive Density under Leave One Out (ELPD-LOO) metric for time series. 
Formally, assume observations $y_1, \dots, y_n$ are independent given parameters $\theta$. Let $p(\theta)$ be a prior distribution on the model parameters that forms a posterior distribution $p(\theta \mid y)$ and posterior predictive $p(\tilde y \mid y) = \int p(\tilde y \mid \theta)p(\theta \mid y)\, d\theta$. We measure predictive accuracy using the expected log pointwise predictive density:
\begin{equation}
\setstretch{1.2}
    \mathrm{ELPD-LOO} = \sum_{i=1}^n \int p_t(\tilde y_i)\, \log p(\tilde y_i \mid y)\, d\tilde y_i
\end{equation}
where $p_t(\tilde{y})$ denotes the true data-generating distribution that we use LOO to approximate. ELPD measures predictive accuracy on a log-probability scale, with higher values indicating better fit. While unbounded below, its absolute value is less important than differences between models, which quantify relative predictive performance. Due to the computational cost of running ELPD-LOO over the full time series, we apply a sliding-window sampling procedure to create a representative subsample (25\%) of the signal. If a sample's ELPD is more than 4 standard deviations from the mean for its agent backbone and method, we flag it as an outlier and re-fit it on the full time series, dividing the ELPD by 4 to normalize (since it is the sum of point-wise calculations). This affects 10 of 630 samples (9 BoxLM, 1 PyVision). All time-series results in the paper use the re-fit values in order to minimize the impact of sub-sampling at 25\%. 

We consider additional metrics for both distribution fitting and time-series modeling in Appendix~\ref{app:alterative_evals}. Results under alternative metrics do not change the rank of the best performing agentic systems.

\subsection{Dataset Setup and Statistics}
\vspace{-0.3em}
Let $\bar{D} = \{D_{1},\ldots,D_{n}\}$ be a collection of datasets where each $D_{i}$ represents a set of $n \in [600, 1500]$ points randomly sampled from a ground-truth probability distribution or time-series signal $D_{\text{True}_{i}}$, whose parameters are drawn uniformly from fixed ranges (see Appendix~\ref{app:data_details}). For each domain and difficulty split we generate 50 (\Easy{}/\Astro{}) or 100 (\Hard{}) such datasets, evaluated against the known ground-truth model. In total, our benchmark consists of 200 discrete probability distributions and 200 time series. 

\section{Experimental Setup}\label{sec:impl}

\subsection{\method{} Implementation Details}
\vspace{-0.3em}

We use the following models as backbones for our \method{}: GPT-5.4-mini, Claude Sonnet 4.6, and Kimi K2.5. For GPT-5.4-mini the reasoning effort is set to ``low'' and for Claude Sonnet 4.6 we select the minimal number of reasoning tokens (1024) due to the iterative nature of our task and exploding token lengths. For all experiments run with \method{} we set the maximum number of iterations  $N=5$, number of proposals per step $p=3$, and we select $R$ to be the Akaike Information Criterion (AIC) as it balances model complexity with fit and does not need a ground truth distribution. For the initial plot, we use a histogram for distributions and simple line plot for time series (See Figure~\ref{fig:data_ex}). 

\vspace{-0.3em}
\paragraph{Expert Visual Tools}
In addition to dynamically generating tools, we can also equip \method{} with tools designed by an expert statistician as an \emph{oracle} experiment.
The expert toolkit remains unaltered during the course of exploration
and
consists of general-purpose functions that target key questions expert statisticians use when fitting complex mixture distributions or noisy time-series. For distribution fitting, the static toolkit includes tools for visualizing QQ plots, tail transform diagnostics, probability plots, and running a Gaussian-Mixture-Model based segmentation with moment estimation, enabling agents to systematically diagnose location, scale, tail, and mixture structure miscalculations. For time series, the expert toolkit includes tools for comparing model fit against actuals, diagnosing residuals, and testing temporal independence via autocorrelation plots and Ljung-Box statistics, together enabling agents to detect underfitting, missed periodicity, and unmodeled dynamics. A complete description of tools is available in Appendix~\ref{app:tools}.

\subsection{Baselines}
\vspace{-0.3em}

We detail exact prompts of all baselines in Appendix~\ref{app:vesta_prompts}. For fair comparison with \method{}, we set the maximum number of iterations $N=5$ for the BoxLM and PyVision baselines. For Claude Code, Codex, and OpenCode, we provide a computational budget that matches the ``worst case'' of \method{}. 
\vspace{-0.3em}
\paragraph{BoxLM.} BoxLM \cite{box_lm} is inspired by the principle of Box's Loop where a language model proposes statistical models using PyMC code and then critiques the probabilistic program over $N$ rounds of iteration. The LLM proposes $M$ models per round and scores each of them using ELPD-LOO, selecting the top $p$ as candidates for the next round. The critic LM then synthesizes the fitted models, their scores, and posterior predictive summary statistics into a natural language hypothesis in order to update the model proposals in the next round of feedback. We minimally adapt BoxLM prompts to better reflect our task goals. 

\vspace{-0.3em}
\paragraph{PyVision.} PyVision \cite{pyvision} is a multi-turn framework designed to enhance multimodal reasoning by allowing VLMs to generate and execute Python code during inference. A VLM receives an input, generates Python code, and executes it in an isolated runtime, with the resulting output fed back into the VLM's context to refine reasoning over multiple turns. The iteration continues until the VLM decides to provide a final answer. We adapt PyVision to propose and fit PyMC models by providing the same intent and PyMC coding guidelines that Visual Exploration Agent receives, keeping the multi-turn execution loop intact.

\vspace{-0.3em}
\paragraph{Agentic Coding Harnesses.} To test whether a general-purpose coding agent can match VESTA's structured exploration, we run each backbone model in its native coding harness. Namely, we use Claude Code for Claude Sonnet 4.6, Codex for GPT-5.4-mini, and OpenCode for Kimi-K2.5. A new session is initialized for each dataset, with code executed in a sandboxed directory containing the data, the initial plot, shell access, standard file read/write/search tools, and a Python environment with PyMC, NumPy, pandas, SciPy, and matplotlib. For each session, we set a computational budget matching VESTA, i.e. stopping the run after proposal and fit of 18 PyMC models or making 54 LLM calls, whichever comes first, and then selecting the best-performing model.

\section{Results and Analysis on \data{}}
\label{sec:results}

\begin{figure}
    \centering
    \includegraphics[width=0.9\linewidth]{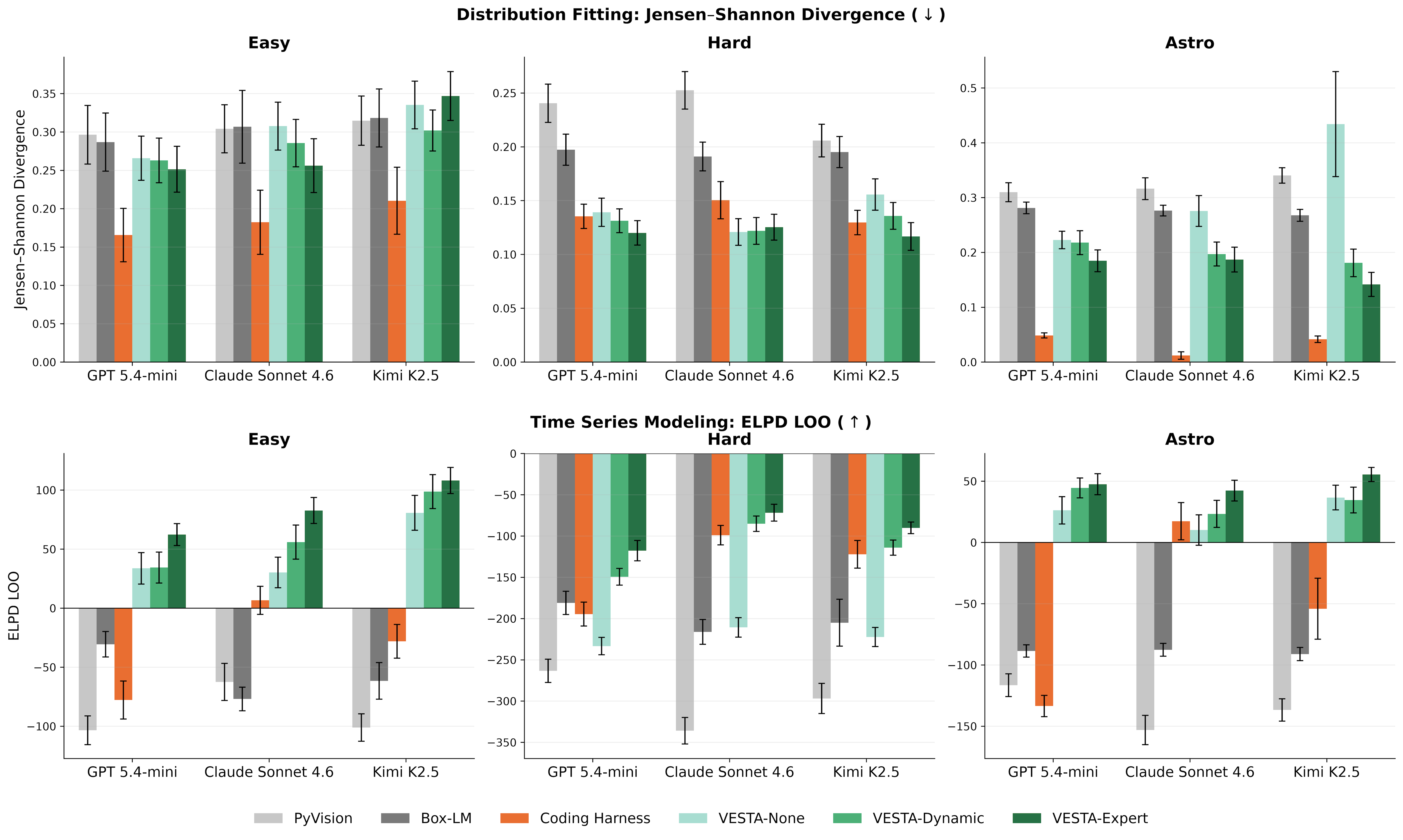}
    \caption{\textbf{[Top]} Average Jensen-Shannon divergence ($\downarrow$ better) between the ground-truth distribution and the probability density function of the proposed \textsc{PyMC} model on the Distribution Fitting task of \data{}.
\textbf{[Bottom]} Average ELPD-LOO ($\uparrow$ better) for the Time Series Modeling task of \data{}, computed via leave-one-out cross-validation. Error bars denote $\pm$1 standard error of the mean.}
    \label{fig:js_bar}
    \vspace{-1em}
\end{figure}

\paragraph{Distribution Fitting Results} 
Figure~\ref{fig:js_bar} reports the Jensen-Shannon Divergence between the ground truth distribution and the probability density function induced by agent-written PyMC code. \method{} equipped with tool use consistently outperforms both the PyVision and BoxLM baselines on the distribution fitting tasks in \data{}.  Coding harnesses (Codex, Claude Code, OpenCode) outperform \method{} on the \protect\Easy{} and \protect\Astro{} splits, whereas \method{} performs better on the \protect\Hard{} mixtures. Inspection of agent traces suggests that coding agents use established numerical fitting libraries to explore popular distributions before submitting PyMC models. Such workflows are effective for single-family problems. Mixture fitting, however, introduces additional choices about component families, weights, and initialization, making effective diagnosis of the current model’s discrepancies more consequential. \method{}'s structured, model-conditioned feedback is particularly useful for such compositional modeling.

When provided with an expert toolkit, \method{} achieves the strongest overall performance, establishing a reference configuration for what tool discovery pipelines can attain. Crucially, \method{} with dynamic tool generation approaches this reference without access to any expert-written tools, producing distributions that far more closely reflect ground truth structure than the no-tool baseline. This gap demonstrates that \method{} is an effective tool discovery method.

\vspace{-0.3em}
\paragraph{Time Series Results} In contrast to Distribution Fitting, \method{} with either dynamic or expert toolkits consistently outperforms PyVision, BoxLM and all coding harnesses, across all three Time Series data splits. \method{} equipped with the expert toolkit has the strongest performance, followed closely by \method{} with the dynamic toolkit. We find that current baselines especially struggle to effectively model time series in \data{}. 

\subsection{Impact of Tool Persistence and Generation in \method{}}
\label{sec:ablation-persistence}

To test the impact of tool accumulation, we sample 100 instances from the \protect\Hard{} distribution-fitting split and evenly partition them into 50 tool discovery and 50 test examples. In the tool discovery phase, \method{} fits PyMC models to the 50 tool discovery examples sequentially with tool accumulation enabled. Each run is initialized with all tools generated in prior runs, and the model either selects an existing tool or writes a new one. We find that accumulation converges quickly, with the registry containing only 4 unique tools after 50 runs. We call the registry after the tool discovery phase the \textit{accumulated toolkit}.
For all \method{} variants, we use a Claude Sonnet 4.6 backbone and report mean Jensen-Shannon divergence on the same 50 test examples. We evaluate four new variants of \method{}. In \method{}-AccOnly, we initialize with the accumulated toolkit and do not allow further tool generation, while in \method{}-Acc+Gen we also allow new tools to be created on the test set. \method{}-Expert+Gen starts with the expert toolkit but allows new tools to be created on the test set. In \method{}-Ephemeral, we begin with an empty toolkit and allow generation, but no tools are stored. We benchmark these against \method{}-Expert and \method{}-Dynamic on the same test set. Table~\ref{tab:persistence-ablation} summarizes each condition and its results.

\begin{table}[h]
    \centering
    \caption{Tool persistence ablation on 50 held-out Hard distribution-fitting examples (Claude Sonnet 4.6). Columns indicate the starting toolkit, whether generation is allowed at test time, and whether generated tools persist across test runs. Mean JS divergence, lower is better.}
    \label{tab:persistence-ablation}
    \resizebox{0.85\linewidth}{!}{%
    \begin{tabular}{lcccc}
        \toprule
        \hline
        Condition & Starting toolkit & Generation & Persistence & Mean JSD \\
        \midrule
        \method{}-AccOnly     & Accumulated toolkit & \xmark & ---    & 0.123 \\
        \method{}-Acc+Gen     & Accumulated toolkit & \cmark & \cmark & 0.115 \\
        \method{}-Expert+Gen  & Expert toolkit      & \cmark & \cmark & \textbf{0.106} \\
        \method{}-Ephemeral   & Empty               & \cmark & \xmark & 0.120 \\
        \midrule
        \method{}-Expert      & Expert toolkit      & \xmark & ---    & 0.129 \\
        \method{}-Dynamic     & Empty               & \cmark & \cmark & 0.124 \\
        \hline
        \bottomrule
    \end{tabular}}
\end{table}

The results shown in Table~\ref{tab:persistence-ablation} indicate that tool generation is a particularly important component of \method{}. In all cases, we see that the ability to generate new tools on unseen examples improves performance.  Generation allows \method{} to construct diagnostics targeted at the specific discrepancies between the current model and the data, which we hypothesize is the reason that this improvement is consistent for all starting toolkit initializations. The gain from \method{}-Expert to \method{}-Expert+Gen further suggests that generated tools complement expert-created tools rather than reproduce them directly. Finally, the small advantage of \method{}-Ephemeral over \method{}-Dynamic is consistent with a growing registry biasing the agent toward reusing existing tools over constructing more targeted ones, although the difference is small and may not be significant at this sample size.

\subsection{Analysis of Generated Tools}
\label{sec:tools}
\vspace{-0.1em}

\begin{wrapfigure}{r}{0.53\textwidth}
  \vspace{-2em}
  \centering
  \includegraphics[width=0.5\textwidth]{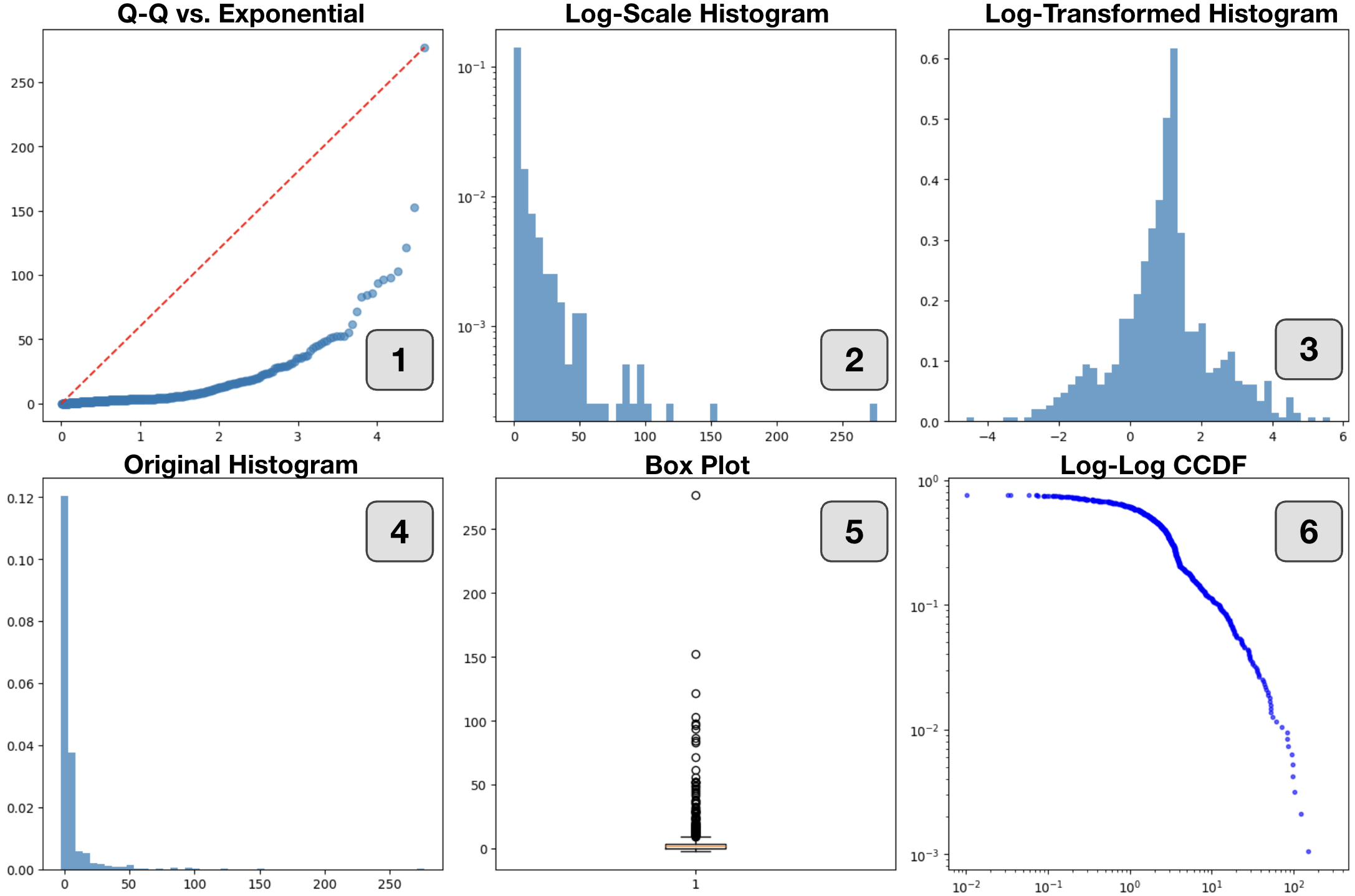}
  \caption{Example of a \method{}-generated tool. This multi-panel output is fed back into \method{} to generate better hypotheses. Panel titles are enlarged for clarity and panel numbers are added manually.}
  \label{fig:tool_ex}
\end{wrapfigure}
\paragraph{Case Study}
Figure~\ref{fig:tool_ex} shows an example of a \method{}-written tool designed to analyze a right-tailed distribution. The tool generated a 6-panel visualization. The original histogram of the input distribution is in \textbf{Panel 4}.

Each of the remaining subplots are designed to test particular hypotheses about this skewed distribution. \textbf{Panel 1} shows a Q-Q plot testing whether the current proposed exponential-distribution-based model is appropriate for this data, shown by whether points fall along the reference line. 
\textbf{Panel 2} plots a histogram of the raw data on a log-scaled $x$-axis, which compresses the right tail and spreads out the lower values, making it easier to assess the shape of the distribution. \textbf{Panel 3} plots a histogram of the log-transformed data, meaning each observation is logged before being binned. If this histogram appears bell-shaped and symmetric then a log-normal family is appropriate for the original data. \textbf{Panel 5} is a box plot and 
\textbf{Panel 6} plots the complementary cumulative distribution function (CCDF) of the data in log-log space to see if the data follows a power-law distribution. 
This multi-panel visualization is a very powerful ingredient for \method{}; one limitation is that VLMs sometimes exhibit a perceptual ``blindness'' \cite{rudman-etal-2025-forgotten,tong2024eyeswideshutexploring} and fail to analyze these complex visualizations, a technical limitation that if addressed, will likely yield further performance gains.
\vspace{-0.3em}
\paragraph{Generated vs.\ Expert Tools}
We analyze the tools generated by \method{} and PyVision using Claude Sonnet 4.6, 
and compare them to the expert-written toolkit. 
We find that \method{} and PyVision often independently ``discover'' tools present in the expert toolkit and compose them together to create a single, powerful tool for analysis.
In contrast, expert tools are atomic by design, often testing at most two hypotheses. 
For example, a single tool generated by \method{} composes an average of 3.5 tools present in the expert toolkit, compared to 1.9 tools for PyVision.  

Table~\ref{tab:tool_results} quantifies this analysis at scale. We calculate the percent of times a generated tool contains a sub-function present in the expert toolkit. First, we find that both methods discover each function type in the toolkit at least once. \method{} recovers a much larger percentage of expert tools than PyVision.

Second, we find that generated tools depend on the unique specifications of the task. For distribution fitting, the \protect\Astro{} split contains Initial Mass Functions (IMFs) which are exclusively power law or mixtures of power law distributions. In order to visually distinguish these distributions, applying a log or log-log transform is required. Accordingly, for \method{} the percent of generated tools using a ``tail transform'' function increases from 55.4\% on the \protect\Easy{} split to 91.8\% on the \protect\Astro{} split. 

\begin{table}
    \centering
    \footnotesize
    \setlength{\tabcolsep}{2pt}
    \caption{Expert tool coverage (\%) across difficulty levels for \method{} and PyVision using Claude Sonnet 4.6. Each value is the percentage of generated tools that implement one of the expert tools.}
    \label{tab:tool_results}
    \begin{tabular}{l rr | rr | rr}
      \toprule \hline
      & \multicolumn{2}{c|}{\protect\Easy{}}
      & \multicolumn{2}{c|}{\protect\Hard{}}
      & \multicolumn{2}{c}{\protect\Astro{}} \\
      \cmidrule(lr){2-3} \cmidrule(lr){4-5} \cmidrule(lr){6-7}
      \textbf{Expert Tool}
        & \textbf{VESTA} & \textbf{PyVision}
        & \textbf{VESTA} & \textbf{PyVision}
        & \textbf{VESTA} & \textbf{PyVision} \\
      \hline
      \multicolumn{7}{l}{\textit{Distributions}} \\
      \hline
      Calculate Moments        & \textbf{98.2} & 72.7          & \textbf{94.6} & 74.4          & \textbf{98.4} & 88.0 \\
      Histogram                & \textbf{98.2} & 14.7          & \textbf{99.2} & 22.8          & \textbf{98.4} &  2.7 \\
      Tails Transforms         & \textbf{55.4} & 44.7          & \textbf{56.2} & 36.7          & \textbf{91.8} & 65.3 \\
      QQ Plot                  & \textbf{60.7} &  4.0          & \textbf{36.2} &  1.1          & \textbf{23.0} &  0.7 \\
      Segment \& CM.           & \textbf{23.2} &  9.3          & \textbf{56.2} & 38.3          & \textbf{14.8} &  0.0 \\
      Probability Plot         & 26.8          & \textbf{30.7} & \textbf{26.9} & 22.8          & \textbf{39.3} & 32.0 \\
      \hline
      \multicolumn{7}{l}{\textit{Time Series}} \\
      \hline
      Residuals ACF Plot                & \textbf{93.8} & 49.3          & \textbf{84.1} & 41.5          & \textbf{97.1} & 50.7 \\
      Fit vs.\ Actuals                  & \textbf{85.2} & 57.3          & \textbf{77.9} & 52.4          & \textbf{67.6} & 56.0 \\
      Instantaneous Freq. Plot      & \textbf{88.9} & 61.3          & \textbf{59.3} & 42.4          & \textbf{82.4} & 58.0 \\
      Fit vs.\ Actuals w/ Res.\ Dist.   & \textbf{76.5} &  7.3          & \textbf{72.4} &  4.2          & \textbf{58.8} &  0.0 \\
      Residuals ACF Score               & 14.8          & \textbf{32.0} &  9.7          & \textbf{18.2} & 13.2          & \textbf{36.7} \\
      \hline \bottomrule 
    \end{tabular}
\end{table}

\paragraph{\method{} Created Tools Beyond the Expert Toolkit} Not only does \method{} discover expert-written tools, but it creates tools that further strengthen analysis. 
\begin{wraptable}{r}{0.5\linewidth}
    \centering
    \footnotesize
    \setlength{\tabcolsep}{4pt}
    \caption{For the distribution fitting task, the percentage of \method{}-generated tools that invoke each function as a subroutine. Only functions not present in the expert toolkit are shown.}
    \label{tab:new_tools}
    \begin{tabular}{l rrr}
      \hline\hline
      \textbf{Function}
        & \textbf{\protect\Easy{}}
        & \textbf{\protect\Hard{}}
        & \textbf{\protect\Astro{}} \\
      \hline
      Diagnostic Fit Checks  & \textbf{100.0} & 96.9          & \textbf{100.0} \\
      Info.\ Criteria        & \textbf{82.1}  & 73.8          & 65.6 \\
      Mean Excess Plot       & 5.4            & 7.7           & \textbf{36.1} \\
      Hill Estimator         & 0.0            & 1.5           & \textbf{6.6}  \\
      Box--Cox               & \textbf{1.8}   & 0.0           & 0.0 \\
      D'Agostino Normaltest  & 0.0            & \textbf{0.8}  & 0.0 \\
      Shapiro--Wilk          & 0.0            & \textbf{0.8}  & 0.0 \\
      \hline\hline
    \end{tabular}
\end{wraptable}
Table~\ref{tab:new_tools} shows the percentage of \method{}-generated tools that contain a function not present in the expert toolkit. We find that nearly all generated tools contain some form of ``diagnostic check'' where the tool fits one or more candidate distributions against the input data to assess which distribution best models the data. This is often accompanied by a formal ranking by calculating different ``information criteria'' metrics such as AIC or BIC. We find the next most common functions (mean-excess and Hill estimator) are predominantly called on the \protect\Astro{} split and are designed to diagnose heavy-tailed behavior. The mean-excess plot exploits linearity of $\mathbb{E}[X - u \mid X > u]$ in the threshold $u$ to identify a Generalized-Pareto-like tail and locate where it begins, while the Hill estimator provides a direct estimate of the tail index $\alpha$ from the largest order statistics. Their focus on \protect\Astro{} suggests that \method{} correctly identifies Initial Mass Functions as tail-dominated distributions and tailors its tools to target these features. The remaining tools (Box–Cox, D'Agostino's normality test, and Shapiro–Wilk) are all related to Gaussian distributions, either testing for normality or transforming data to be normally distributed. A detailed description of each tool, as well as an analysis of novel time-series tools is presented in Appendix~\ref{app:detailed_tool_analysis}.

\section{Conclusion} 
We introduce \method{}, a pipeline that uses VLMs to propose statistical models and test hypotheses by way of visual diagnostic tools, producing models more accurately reflecting the underlying data structure. To support this evaluation, we introduce \data{}, a benchmark of distribution fitting and time series modeling tasks of increasing difficulty, including real-world astronomy problems (\protect\Astro{}). 
Our evaluation finds that \method{}'s dynamic tool generation both independently recovers expert tools and composes them into sophisticated diagnostics to test multiple hypotheses simultaneously. While expert tools achieve the strongest performance, \method{} outperforms existing baselines on complex distribution fitting and time-series modeling tasks, demonstrating that autonomous discovery of visual diagnostic tools can lead to fitting higher-quality statistical models.

\section*{Acknowledgments}

This work was supported by the NSF under Cooperative Agreement 2421782 and the Simons Foundation grant MPS-AI-00010515 awarded to the NSF-Simons AI Institute for Cosmic Origins - CosmicAI, \url{https://www.cosmicai.org/}. It was also partially supported by NSF grants IIS-2145280, IIS-2433071, and CCF-2403036. We would also like to thank Stéphanie Juneau from CosmicAI / NOIRLab for her help and feedback on this work.

\bibliographystyle{plainnat}
\bibliography{bib}

@misc{llm_zero_shot_ts,
      title={Large Language Models Are Zero-Shot Time Series Forecasters}, 
      author={Nate Gruver and Marc Finzi and Shikai Qiu and Andrew Gordon Wilson},
      year={2024},
      eprint={2310.07820},
      archivePrefix={arXiv},
      primaryClass={cs.LG},
      url={https://arxiv.org/abs/2310.07820}, 
}

@misc{transformer_ts_survey,
      title={Transformers in Time Series: A Survey}, 
      author={Qingsong Wen and Tian Zhou and Chaoli Zhang and Weiqi Chen and Ziqing Ma and Junchi Yan and Liang Sun},
      year={2023},
      eprint={2202.07125},
      archivePrefix={arXiv},
      primaryClass={cs.LG},
      url={https://arxiv.org/abs/2202.07125}, 
}

@misc{time_series_transformer,
      title={A Transformer-based Framework for Multivariate Time Series Representation Learning}, 
      author={George Zerveas and Srideepika Jayaraman and Dhaval Patel and Anuradha Bhamidipaty and Carsten Eickhoff},
      year={2020},
      eprint={2010.02803},
      archivePrefix={arXiv},
      primaryClass={cs.LG},
      url={https://arxiv.org/abs/2010.02803}, 
}

@misc{time_vlm,
      title={Time-VLM: Exploring Multimodal Vision-Language Models for Augmented Time Series Forecasting}, 
      author={Siru Zhong and Weilin Ruan and Ming Jin and Huan Li and Qingsong Wen and Yuxuan Liang},
      year={2025},
      eprint={2502.04395},
      archivePrefix={arXiv},
      primaryClass={cs.CV},
      url={https://arxiv.org/abs/2502.04395}, 
}

@misc{vlm_ts_survey,
      title={Harnessing Vision Models for Time Series Analysis: A Survey}, 
      author={Jingchao Ni and Ziming Zhao and ChengAo Shen and Hanghang Tong and Dongjin Song and Wei Cheng and Dongsheng Luo and Haifeng Chen},
      year={2025},
      eprint={2502.08869},
      archivePrefix={arXiv},
      primaryClass={cs.LG},
      url={https://arxiv.org/abs/2502.08869}, 
}

@article{llm_statisticians,
   title={A Survey on Large Language Model-based Agents for Statistics and Data Science},
   ISSN={1537-2731},
   url={http://dx.doi.org/10.1080/00031305.2025.2561140},
   DOI={10.1080/00031305.2025.2561140},
   journal={The American Statistician},
   publisher={Informa UK Limited},
   author={Sun, Maojun and Han, Ruijian and Jiang, Binyan and Qi, Houduo and Sun, Defeng and Yuan, Yancheng and Huang, Jian},
   year={2025},
   month=Oct, pages={1–14} }

@misc{are_llms_good_statisticians,
      title={Are Large Language Models Good Statisticians?}, 
      author={Yizhang Zhu and Shiyin Du and Boyan Li and Yuyu Luo and Nan Tang},
      year={2024},
      eprint={2406.07815},
      archivePrefix={arXiv},
      primaryClass={cs.CL},
      url={https://arxiv.org/abs/2406.07815}, 
}

@misc{llm_stat_overview,
      title={An Overview of Large Language Models for Statisticians}, 
      author={Wenlong Ji and Weizhe Yuan and Emily Getzen and Kyunghyun Cho and Michael I. Jordan and Song Mei and Jason E Weston and Weijie J. Su and Jing Xu and Linjun Zhang},
      year={2025},
      eprint={2502.17814},
      archivePrefix={arXiv},
      primaryClass={stat.ML},
      url={https://arxiv.org/abs/2502.17814}, 
}

@misc{boxing_gym,
      title={BoxingGym: Benchmarking Progress in Automated Experimental Design and Model Discovery}, 
      author={Kanishk Gandhi and Michael Y. Li and Lyle Goodyear and Agam Bhatia and Louise Li and Aditi Bhaskar and Mohammed Zaman and Noah D. Goodman},
      year={2025},
      eprint={2501.01540},
      archivePrefix={arXiv},
      primaryClass={cs.LG},
      url={https://arxiv.org/abs/2501.01540}, 
}

@misc{box_lm,
      title={Automated Statistical Model Discovery with Language Models}, 
      author={Michael Y. Li and Emily B. Fox and Noah D. Goodman},
      year={2024},
      eprint={2402.17879},
      archivePrefix={arXiv},
      primaryClass={cs.LG},
      url={https://arxiv.org/abs/2402.17879}, 
}

@misc{multimodal_amd,
      title={Automated Model Discovery via Multi-modal \& Multi-step Pipeline}, 
      author={Lee Jung-Mok and Nam Hyeon-Woo and Moon Ye-Bin and Junhyun Nam and Tae-Hyun Oh},
      year={2025},
      eprint={2509.25946},
      archivePrefix={arXiv},
      primaryClass={cs.AI},
      url={https://arxiv.org/abs/2509.25946}, 
}

@misc{critical,
      title={CriticAL: Critic Automation with Language Models}, 
      author={Michael Y. Li and Vivek Vajipey and Noah D. Goodman and Emily B. Fox},
      year={2024},
      eprint={2411.06590},
      archivePrefix={arXiv},
      primaryClass={cs.LG},
      url={https://arxiv.org/abs/2411.06590}, 
}

@article{probabilistic_grammars,
  author       = {Jure Brence and
                  Ljupco Todorovski and
                  Saso Dzeroski},
  title        = {Probabilistic Grammars for Equation Discovery},
  journal      = {CoRR},
  volume       = {abs/2012.00428},
  year         = {2020},
  url          = {https://arxiv.org/abs/2012.00428},
  eprinttype   = {arXiv},
  eprint       = {2012.00428},
  bibsource    = {dblp computer science bibliography, https://dblp.org}
}

@misc{kernel_search,
      title={Structure Discovery in Nonparametric Regression through Compositional Kernel Search}, 
      author={David Duvenaud and James Robert Lloyd and Roger Grosse and Joshua B. Tenenbaum and Zoubin Ghahramani},
      year={2013},
      eprint={1302.4922},
      archivePrefix={arXiv},
      primaryClass={stat.ML},
      url={https://arxiv.org/abs/1302.4922}, 
}

@article{amd_non_linear_dynamics,
  author    = {Bongard, Josh and Lipson, Hod},
  title     = {Automated reverse engineering of nonlinear dynamical systems},
  journal   = {Proceedings of the National Academy of Sciences of the United States of America},
  year      = {2007},
  volume    = {104},
  number    = {24},
  pages     = {9943--9948},
  month     = {Jun},
  doi       = {10.1073/pnas.0609476104},
  pmid      = {17553966},
  pmcid     = {PMC1891254},
  publisher = {National Academy of Sciences},
}

@misc{llm_processes,
      title={LLM Processes: Numerical Predictive Distributions Conditioned on Natural Language}, 
      author={James Requeima and John Bronskill and Dami Choi and Richard E. Turner and David Duvenaud},
      year={2024},
      eprint={2405.12856},
      archivePrefix={arXiv},
      primaryClass={stat.ML},
      url={https://arxiv.org/abs/2405.12856}, 
}

@misc{pyvision,
      title={PyVision: Agentic Vision with Dynamic Tooling}, 
      author={Shitian Zhao and Haoquan Zhang and Shaoheng Lin and Ming Li and Qilong Wu and Kaipeng Zhang and Chen Wei},
      year={2025},
      eprint={2507.07998},
      archivePrefix={arXiv},
      primaryClass={cs.CL},
      url={https://arxiv.org/abs/2507.07998}, 
}

@misc{vipact,
      title={VipAct: Visual-Perception Enhancement via Specialized VLM Agent Collaboration and Tool-use}, 
      author={Zhehao Zhang and Ryan Rossi and Tong Yu and Franck Dernoncourt and Ruiyi Zhang and Jiuxiang Gu and Sungchul Kim and Xiang Chen and Zichao Wang and Nedim Lipka},
      year={2025},
      eprint={2410.16400},
      archivePrefix={arXiv},
      primaryClass={cs.CL},
      url={https://arxiv.org/abs/2410.16400}, 
}

@misc{visual_sketchpad,
      title={Visual Sketchpad: Sketching as a Visual Chain of Thought for Multimodal Language Models}, 
      author={Yushi Hu and Weijia Shi and Xingyu Fu and Dan Roth and Mari Ostendorf and Luke Zettlemoyer and Noah A Smith and Ranjay Krishna},
      year={2024},
      eprint={2406.09403},
      archivePrefix={arXiv},
      primaryClass={cs.CV},
      url={https://arxiv.org/abs/2406.09403}, 
}

@misc{image_of_thought,
      title={Image-of-Thought Prompting for Visual Reasoning Refinement in Multimodal Large Language Models}, 
      author={Qiji Zhou and Ruochen Zhou and Zike Hu and Panzhong Lu and Siyang Gao and Yue Zhang},
      year={2024},
      eprint={2405.13872},
      archivePrefix={arXiv},
      primaryClass={cs.AI},
      url={https://arxiv.org/abs/2405.13872}, 
}

@misc{visual_chagpt,
      title={Visual ChatGPT: Talking, Drawing and Editing with Visual Foundation Models}, 
      author={Chenfei Wu and Shengming Yin and Weizhen Qi and Xiaodong Wang and Zecheng Tang and Nan Duan},
      year={2023},
      eprint={2303.04671},
      archivePrefix={arXiv},
      primaryClass={cs.CV},
      url={https://arxiv.org/abs/2303.04671}, 
}

@misc{visual_programming,
      title={Visual Programming: Compositional visual reasoning without training}, 
      author={Tanmay Gupta and Aniruddha Kembhavi},
      year={2022},
      eprint={2211.11559},
      archivePrefix={arXiv},
      primaryClass={cs.CV},
      url={https://arxiv.org/abs/2211.11559}, 
}

@misc{viper_gpt,
      title={ViperGPT: Visual Inference via Python Execution for Reasoning}, 
      author={Dídac Surís and Sachit Menon and Carl Vondrick},
      year={2023},
      eprint={2303.08128},
      archivePrefix={arXiv},
      primaryClass={cs.CV},
      url={https://arxiv.org/abs/2303.08128}, 
}

@misc{visual_perception_w_tools,
      title={Reinforced Visual Perception with Tools}, 
      author={Zetong Zhou and Dongping Chen and Zixian Ma and Zhihan Hu and Mingyang Fu and Sinan Wang and Yao Wan and Zhou Zhao and Ranjay Krishna},
      year={2025},
      eprint={2509.01656},
      archivePrefix={arXiv},
      primaryClass={cs.CV},
      url={https://arxiv.org/abs/2509.01656}, 
}

@misc{vis_mem,
      title={VisMem: Latent Vision Memory Unlocks Potential of Vision-Language Models}, 
      author={Xinlei Yu and Chengming Xu and Guibin Zhang and Zhangquan Chen and Yudong Zhang and Yongbo He and Peng-Tao Jiang and Jiangning Zhang and Xiaobin Hu and Shuicheng Yan},
      year={2026},
      eprint={2511.11007},
      archivePrefix={arXiv},
      primaryClass={cs.CV},
      url={https://arxiv.org/abs/2511.11007}, 
}

@misc{skywork,
      title={Skywork-R1V4: Toward Agentic Multimodal Intelligence through Interleaved Thinking with Images and DeepResearch}, 
      author={Yifan Zhang and Liang Hu and Haofeng Sun and Peiyu Wang and Yichen Wei and Shukang Yin and Jiangbo Pei and Wei Shen and Peng Xia and Yi Peng and Tianyidan Xie and Eric Li and Yang Liu and Xuchen Song and Yahui Zhou},
      year={2025},
      eprint={2512.02395},
      archivePrefix={arXiv},
      primaryClass={cs.CV},
      url={https://arxiv.org/abs/2512.02395}, 
}

@misc{deep_eyes,
      title={DeepEyesV2: Toward Agentic Multimodal Model}, 
      author={Jack Hong and Chenxiao Zhao and ChengLin Zhu and Weiheng Lu and Guohai Xu and Xing Yu},
      year={2026},
      eprint={2511.05271},
      archivePrefix={arXiv},
      primaryClass={cs.CV},
      url={https://arxiv.org/abs/2511.05271}, 
}

@misc{act_wisely,
      title={Act Wisely: Cultivating Meta-Cognitive Tool Use in Agentic Multimodal Models}, 
      author={Shilin Yan and Jintao Tong and Hongwei Xue and Xiaojun Tang and Yangyang Wang and Kunyu Shi and Guannan Zhang and Ruixuan Li and Yixiong Zou},
      year={2026},
      eprint={2604.08545},
      archivePrefix={arXiv},
      primaryClass={cs.CV},
      url={https://arxiv.org/abs/2604.08545}, 
}

@misc{codev,
      title={CodeV: Code with Images for Faithful Visual Reasoning via Tool-Aware Policy Optimization}, 
      author={Xinhai Hou and Shaoyuan Xu and Manan Biyani and Moyan Li and Jia Liu and Todd C. Hollon and Bryan Wang},
      year={2026},
      eprint={2511.19661},
      archivePrefix={arXiv},
      primaryClass={cs.CV},
      url={https://arxiv.org/abs/2511.19661}, 
}

@misc{latm,
      title={Large Language Models as Tool Makers}, 
      author={Tianle Cai and Xuezhi Wang and Tengyu Ma and Xinyun Chen and Denny Zhou},
      year={2024},
      eprint={2305.17126},
      archivePrefix={arXiv},
      primaryClass={cs.LG},
      url={https://arxiv.org/abs/2305.17126}, 
}

@misc{creator,
      title={CREATOR: Tool Creation for Disentangling Abstract and Concrete Reasoning of Large Language Models}, 
      author={Cheng Qian and Chi Han and Yi R. Fung and Yujia Qin and Zhiyuan Liu and Heng Ji},
      year={2024},
      eprint={2305.14318},
      archivePrefix={arXiv},
      primaryClass={cs.CL},
      url={https://arxiv.org/abs/2305.14318}, 
}

@misc{tookmaker,
      title={LLM Agents Making Agent Tools}, 
      author={Georg Wölflein and Dyke Ferber and Daniel Truhn and Ognjen Arandjelović and Jakob Nikolas Kather},
      year={2025},
      eprint={2502.11705},
      archivePrefix={arXiv},
      primaryClass={cs.CL},
      url={https://arxiv.org/abs/2502.11705}, 
}

@misc{evoskill,
      title={EvoSkill: Automated Skill Discovery for Multi-Agent Systems}, 
      author={Salaheddin Alzubi and Noah Provenzano and Jaydon Bingham and Weiyuan Chen and Tu Vu},
      year={2026},
      eprint={2603.02766},
      archivePrefix={arXiv},
      primaryClass={cs.AI},
      url={https://arxiv.org/abs/2603.02766}, 
}

@misc{voyager,
      title={Voyager: An Open-Ended Embodied Agent with Large Language Models}, 
      author={Guanzhi Wang and Yuqi Xie and Yunfan Jiang and Ajay Mandlekar and Chaowei Xiao and Yuke Zhu and Linxi Fan and Anima Anandkumar},
      year={2023},
      eprint={2305.16291},
      archivePrefix={arXiv},
      primaryClass={cs.AI},
      url={https://arxiv.org/abs/2305.16291}, 
}

@misc{toolace,
      title={ToolACE-DEV: Self-Improving Tool Learning via Decomposition and EVolution}, 
      author={Xu Huang and Weiwen Liu and Xingshan Zeng and Yuefeng Huang and Xinlong Hao and Yuxian Wang and Yirong Zeng and Chuhan Wu and Yasheng Wang and Ruiming Tang and Defu Lian},
      year={2025},
      eprint={2505.07512},
      archivePrefix={arXiv},
      primaryClass={cs.CL},
      url={https://arxiv.org/abs/2505.07512}, 
}

@misc{evoclaw,
      title={EvoClaw: Evaluating AI Agents on Continuous Software Evolution}, 
      author={Gangda Deng and Zhaoling Chen and Zhongming Yu and Haoyang Fan and Yuhong Liu and Yuxin Yang and Dhruv Parikh and Rajgopal Kannan and Le Cong and Mengdi Wang and Qian Zhang and Viktor Prasanna and Xiangru Tang and Xingyao Wang},
      year={2026},
      eprint={2603.13428},
      archivePrefix={arXiv},
      primaryClass={cs.SE},
      url={https://arxiv.org/abs/2603.13428}, 
}

@misc{sea-agent,
      title={SEAgent: Self-Evolving Computer Use Agent with Autonomous Learning from Experience}, 
      author={Zeyi Sun and Ziyu Liu and Yuhang Zang and Yuhang Cao and Xiaoyi Dong and Tong Wu and Dahua Lin and Jiaqi Wang},
      year={2025},
      eprint={2508.04700},
      archivePrefix={arXiv},
      primaryClass={cs.AI},
      url={https://arxiv.org/abs/2508.04700}, 
}

@misc{tte,
      title={Beyond Static Tools: Test-Time Tool Evolution for Scientific Reasoning}, 
      author={Jiaxuan Lu and Ziyu Kong and Yemin Wang and Rong Fu and Haiyuan Wan and Cheng Yang and Wenjie Lou and Haoran Sun and Lilong Wang and Yankai Jiang and Xiaosong Wang and Xiao Sun and Dongzhan Zhou},
      year={2026},
      eprint={2601.07641},
      archivePrefix={arXiv},
      primaryClass={cs.AI},
      url={https://arxiv.org/abs/2601.07641}, 
}

@misc{rest-gpt,
      title={RestGPT: Connecting Large Language Models with Real-World RESTful APIs}, 
      author={Yifan Song and Weimin Xiong and Dawei Zhu and Wenhao Wu and Han Qian and Mingbo Song and Hailiang Huang and Cheng Li and Ke Wang and Rong Yao and Ye Tian and Sujian Li},
      year={2023},
      eprint={2306.06624},
      archivePrefix={arXiv},
      primaryClass={cs.CL},
      url={https://arxiv.org/abs/2306.06624}, 
}

@misc{open-sage,
      title={OpenSage: Self-programming Agent Generation Engine}, 
      author={Hongwei Li and Zhun Wang and Qinrun Dai and Yuzhou Nie and Jinjun Peng and Ruitong Liu and Jingyang Zhang and Kaijie Zhu and Jingxuan He and Lun Wang and Yangruibo Ding and Yueqi Chen and Wenbo Guo and Dawn Song},
      year={2026},
      eprint={2602.16891},
      archivePrefix={arXiv},
      primaryClass={cs.AI},
      url={https://arxiv.org/abs/2602.16891}, 
}

@misc{dab_step,
      title={DABstep: Data Agent Benchmark for Multi-step Reasoning}, 
      author={Alex Egg and Martin Iglesias Goyanes and Friso Kingma and Andreu Mora and Leandro von Werra and Thomas Wolf},
      year={2025},
      eprint={2506.23719},
      archivePrefix={arXiv},
      primaryClass={cs.LG},
      url={https://arxiv.org/abs/2506.23719}, 
}

@misc{mle_bench,
      title={MLE-bench: Evaluating Machine Learning Agents on Machine Learning Engineering}, 
      author={Jun Shern Chan and Neil Chowdhury and Oliver Jaffe and James Aung and Dane Sherburn and Evan Mays and Giulio Starace and Kevin Liu and Leon Maksin and Tejal Patwardhan and Lilian Weng and Aleksander Mądry},
      year={2025},
      eprint={2410.07095},
      archivePrefix={arXiv},
      primaryClass={cs.CL},
      url={https://arxiv.org/abs/2410.07095}, 
}

@misc{ada_evolve,
      title={AdaEvolve: Adaptive LLM Driven Zeroth-Order Optimization}, 
      author={Mert Cemri and Shubham Agrawal and Akshat Gupta and Shu Liu and Audrey Cheng and Qiuyang Mang and Ashwin Naren and Lutfi Eren Erdogan and Koushik Sen and Matei Zaharia and Alex Dimakis and Ion Stoica},
      year={2026},
      eprint={2602.20133},
      archivePrefix={arXiv},
      primaryClass={cs.NE},
      url={https://arxiv.org/abs/2602.20133}, 
}

@misc{execution_grounded_research,
      title={Towards Execution-Grounded Automated AI Research}, 
      author={Chenglei Si and Zitong Yang and Yejin Choi and Emmanuel Candès and Diyi Yang and Tatsunori Hashimoto},
      year={2026},
      eprint={2601.14525},
      archivePrefix={arXiv},
      primaryClass={cs.CL},
      url={https://arxiv.org/abs/2601.14525}, 
}

@misc{NASA4Vesta,
  author       = {{NASA Science}},
  title        = {Vesta: In Depth},
  howpublished = {\url{https://science.nasa.gov/solar-system/asteroids/4-vesta/}},
  note         = {Accessed: May 2, 2026}
}

@misc{NASADawnMission,
  author       = {{NASA Science}},
  title        = {Dawn Mission Overview},
  howpublished = {\url{https://science.nasa.gov/mission/dawn/}},
  note         = {Accessed: May 2, 2026}
}

@article{annurev-astro-082708-101642,
   author = "Bastian, Nate and Covey, Kevin R. and Meyer, Michael R.",
   title = "A Universal Stellar Initial Mass Function? A Critical Look at Variations", 
   journal= "Annual Review of Astronomy and Astrophysics",
   year = "2010",
   volume = "48",
   number = "Volume 48, 2010",
   pages = "339-389",
   doi = "https://doi.org/10.1146/annurev-astro-082708-101642",
   url = "https://www.annualreviews.org/content/journals/10.1146/annurev-astro-082708-101642",
   publisher = "Annual Reviews",
   issn = "1545-4282",
   type = "Journal Article",
  }

@article{pymc,
  title={PyMC: a modern, and comprehensive probabilistic programming framework in Python},
  author={Abril-Pla, Oriol and Andreani, Virgile and Carroll, Colin and Dong, Larry and Fonnesbeck, Christopher J and Kochurov, Maxim and Kumar, Ravin and Lao, Junpeng and Luhmann, Christian C and Martin, Osvaldo A and others},
  journal={PeerJ Computer Science},
  volume={9},
  pages={e1516},
  year={2023},
  publisher={PeerJ Inc.}
}

@misc{lloyd2014automaticconstructionnaturallanguagedescription,
      title={Automatic Construction and Natural-Language Description of Nonparametric Regression Models}, 
      author={James Robert Lloyd and David Duvenaud and Roger Grosse and Joshua B. Tenenbaum and Zoubin Ghahramani},
      year={2014},
      eprint={1402.4304},
      archivePrefix={arXiv},
      primaryClass={stat.ML},
      url={https://arxiv.org/abs/1402.4304}, 
}

@article{joseph2025astrovisbench,
  title={Astrovisbench: A code benchmark for scientific computing and visualization in astronomy},
  author={Joseph, Sebastian Antony and Husain, Syed Murtaza and Offner, Stella SR and Juneau, St{\~A}{\v{S}}phanie and Torrey, Paul and Bolton, Adam S and Farias, Juan P and Gaffney, Niall and Durrett, Greg and Li, Junyi Jessy},
  journal={arXiv preprint arXiv:2505.20538},
  year={2025}
}

@inbook{imf_survey,
   title={The Origin and Universality of the Stellar Initial Mass Function},
   ISBN={9780816531240},
   url={http://dx.doi.org/10.2458/azu_uapress_9780816531240-ch003},
   DOI={10.2458/azu_uapress_9780816531240-ch003},
   booktitle={Protostars and Planets VI},
   publisher={University of Arizona Press},
   author={Offner, S. S. R. and Clark, P. C. and Hennebelle, P. and Bastian, N. and Bate, M. R. and Hopkins, P. F. and Moreaux, E. and Whitworth, A. P.},
   year={2014} }

@ARTICLE{salpeter,
       author = {{Salpeter}, Edwin E.},
        title = "{The Luminosity Function and Stellar Evolution.}",
      journal = {apj},
         year = 1955,
        month = jan,
       volume = {121},
        pages = {161},
          doi = {10.1086/145971},
       adsurl = {https://ui.adsabs.harvard.edu/abs/1955ApJ...121..161S}
}

@article{Kroupa_2001,
   title={On the variation of the initial mass function},
   volume={322},
   ISSN={1365-2966},
   url={http://dx.doi.org/10.1046/j.1365-8711.2001.04022.x},
   DOI={10.1046/j.1365-8711.2001.04022.x},
   number={2},
   journal={Monthly Notices of the Royal Astronomical Society},
   publisher={Oxford University Press (OUP)},
   author={Kroupa, P.},
   year={2001},
   month=Apr, pages={231–246} }

@article{Chabrier_2003,
   title={Galactic Stellar and Substellar Initial Mass Function},
   volume={115},
   ISSN={1538-3873},
   url={http://dx.doi.org/10.1086/376392},
   DOI={10.1086/376392},
   number={809},
   journal={Publications of the Astronomical Society of the Pacific},
   publisher={IOP Publishing},
   author={Chabrier, Gilles},
   year={2003},
   month="July", pages={763–795} }

@inproceedings{rudman-etal-2025-forgotten,
    title = "Forgotten Polygons: Multimodal Large Language Models are Shape-Blind",
    author = "Rudman, William  and
      Golovanevsky, Michal  and
      Bar, Amir  and
      Palit, Vedant  and
      LeCun, Yann  and
      Eickhoff, Carsten  and
      Singh, Ritambhara",
    editor = "Che, Wanxiang  and
      Nabende, Joyce  and
      Shutova, Ekaterina  and
      Pilehvar, Mohammad Taher",
    booktitle = "Findings of the Association for Computational Linguistics: ACL 2025",
    month = jul,
    year = "2025",
    address = "Vienna, Austria",
    publisher = "Association for Computational Linguistics",
    url = "https://aclanthology.org/2025.findings-acl.620/",
    doi = "10.18653/v1/2025.findings-acl.620",
    pages = "11983--11998",
    ISBN = "979-8-89176-256-5"
}

@misc{tong2024eyeswideshutexploring,
      title={Eyes Wide Shut? Exploring the Visual Shortcomings of Multimodal LLMs}, 
      author={Shengbang Tong and Zhuang Liu and Yuexiang Zhai and Yi Ma and Yann LeCun and Saining Xie},
      year={2024},
      eprint={2401.06209},
      archivePrefix={arXiv},
      primaryClass={cs.CV},
      url={https://arxiv.org/abs/2401.06209}, 
}

@article{speech_signal,
  author  = {Aroudi, Ali and Veisi, Hadi and Sameti, Hossein and Mafakheri, Zahra},
  year    = {2015},
  month   = {12},
  pages   = {1-14},
  title   = {Speech signal modeling using multivariate distributions},
  volume  = {2015},
  journal = {EURASIP Journal on Audio Speech and Music Processing},
  doi     = {10.1186/s13636-015-0078-1}
}

@article{maller2024mixture,
  author  = {Maller, Ross and Resnick, Sidney and Shemehsavar, Soudabeh and Zhao, Muzhi},
  year    = {2024},
  month   = {01},
  pages   = {},
  title   = {Mixture cure model methodology in survival analysis: Some recent results for the one-sample case},
  volume  = {18},
  journal = {Statistics Surveys},
  doi     = {10.1214/24-SS147}
}

@article{ensafi2022timeseries,
  author   = {Ensafi, Yasaman and Hassanzadeh Amin, Saman and Zhang, Guoqing and Shah, Bharat},
  title    = {Time-series forecasting of seasonal items sales using machine learning -- A comparative analysis},
  journal  = {International Journal of Information Management Data Insights},
  volume   = {2},
  number   = {1},
  pages    = {100058},
  year     = {2022},
  issn     = {2667-0968},
  doi      = {10.1016/j.jjimei.2022.100058},
  url      = {https://www.sciencedirect.com/science/article/pii/S2667096822000027}
}

\newpage

\appendix

\section{Author contributions}

William Rudman, Abhishek Divekar, and Kanishk Jain contributed equally, with complementary responsibilities.

\textbf{William Rudman} led the overall research direction, study design, and scientific framing. He developed the initial modeling framework, led distribution-fitting benchmark development, evaluation, and analysis, and coordinated research across modeling domains and with domain experts. He led manuscript preparation and the synthesis of scientific findings.

\textbf{Abhishek Divekar} led the development of VESTA's central dynamic visual tool-creation methodology. He was the principal implementer of the unified framework and co-led computational experimentation and validation across distribution-fitting and time-series tasks, contributing to study design, analysis, and manuscript revision. His work on this paper does not relate to his affiliation with Amazon.

\textbf{Kanishk Jain} led the development of the time-series forecasting methodology and its dedicated modeling framework, including the workflow, dataset development, domain-specific tools, and performance evaluation. He worked with William Rudman and Abhishek Divekar to unify the time-series and distribution-fitting frameworks into a common, production-ready design for final experiments. He generated and analyzed time-series results and contributed to manuscript preparation.

\textbf{Sebastian Joseph} contributed to research discussions and exploratory baseline development.

\textbf{Stella S. R. Offner} provided astrophysical expertise and guidance on scientific applications. \textbf{Matthew Lease} and \textbf{Kyle Mahowald} provided research guidance and feedback on the project's development and scientific framing.

\textbf{Greg Durrett} and \textbf{Junyi Jessy Li} contributed equally to advising and supervision. They guided the formulation of research questions, methodological development, and experimental design. They helped position the work within the relevant literature, interpret findings, and refine the paper's scientific claims, and contributed to manuscript revision.


\section{\method{} Details}
\begin{algorithm}
\caption{Visual Exploration Agents (Detailed)}
\begin{algorithmic}[1]
\Require Data $D$, iteration limit $N$, proposals per iteration $p$, metric $R$, registry $\mathcal{E}$ (initial state: \texttt{generate\_new\_tool} only)
\Ensure $M_{\text{best}}, \theta_{\text{best}}$
\State $P_0 \gets \text{Plot}(D)$
\State $\{M_0^{(j)}\}_{j=1}^{p} \gets \textsc{Propose}(P_0)$ \Comment{Proposal Agent proposes $p$ diverse models}
\State $\theta_0^{(j)} \gets \textsc{StatGen}(D, M_0^{(j)})\ \forall j$ \Comment{Stat-Gen fits all $p$ models}
\State $M_0, \theta_0 \gets \arg\min_{j}\ R(D, M_0^{(j)}(\theta_0^{(j)}))$ \Comment{Select best model under metric $R$}
\State $M_{\text{best}}, \theta_{\text{best}} \gets M_0, \theta_0$
\State $s_0 \gets \textsc{Summarize}(M_{\text{best}}, \theta_{\text{best}})$
\For{$i = 1, \ldots, N$}
    \State $\varepsilon^* \gets \textsc{ToolManager}(s_{i-1}, D, M_{i-1}, \theta_{i-1}, \mathcal{E})$ \Comment{Select or \textit{create} diagnostic tool}
    \State $\mathcal{E} \gets \mathcal{E} \cup \{\varepsilon^*\}$ \Comment{Persist newly created tools to registry}
    \State $T_i \gets \varepsilon^*(D, M_{i-1}, \theta_{i-1})$ \Comment{Execute tool to produce diagnostic}
    \State $\text{desc},\ \{M_i^{(j)}\}_{j=1}^{p} \gets \textsc{Critique}(T_i, s_{0:i-1}, D)$ \Comment{Propose $p$ revised models}
    \If{$\text{desc} = \texttt{COMPLETE}$}
        \State \textbf{break}
    \EndIf
    \State $\theta_i^{(j)} \gets \textsc{StatGen}(D, M_i^{(j)})\ \forall j$ \Comment{Stat-Gen fits all $p$ models}
    \State $M_i, \theta_i \gets \arg\min_{j}\ R(D, M_i^{(j)}(\theta_i^{(j)}))$ \Comment{Select best model under metric $R$}
    \State $M_{\text{best}}, \theta_{\text{best}} \gets \arg\min_{\theta \in \{\theta_i,\, \theta_{\text{best}}\}} R(D, \mathcal{M}(\theta))$
    \State $s_i \gets \textsc{Summarize}(M_i, \theta_i, \varepsilon^*)$
\EndFor
\State \Return $M_{\text{best}}, \theta_{\text{best}}$
\end{algorithmic}
\label{alg:vea_details}
\end{algorithm}

\section{\data{} Details.}
\label{app:data_details}
\subsection{Distribution Fitting}
For each $D_{\text{True}_i}$ in the distribution fitting domain, ground-truth
parameters are drawn independently and uniformly from predefined ranges, with
the family (or mixture of families) depending on the difficulty split. Nine
base families are used throughout: Pareto, Gaussian, Log-normal, Student-$t$,
Exponential, Uniform, Weibull, Cauchy, and Laplace.

\paragraph{Easy split.}
A single distribution family is drawn uniformly at random from the eight
families above. Parameters are then sampled as follows. For the
\emph{Gaussian}, mean $\mu \sim \mathcal{U}(-5, 5)$ and standard deviation
$\sigma \sim \mathcal{U}(0.5, 3)$. For the \emph{Log-normal}, log-mean
$\mu \sim \mathcal{U}(0, 2)$ and log-width $\sigma \sim \mathcal{U}(0.3,
1.5)$. For the \emph{Student-$t$}, location $\mu \sim \mathcal{U}(-5, 5)$,
scale $\sigma \sim \mathcal{U}(0.5, 3)$, and degrees of freedom $\nu \sim
\mathcal{U}(2.5, 10)$. For the \emph{Exponential}, location $\ell \sim
\mathcal{U}(0, 2)$ and scale $\lambda \sim \mathcal{U}(0.5, 3)$. For the
\emph{Uniform}, mean $\mu \sim \mathcal{U}(-5, 5)$ and standard deviation
$\sigma \sim \mathcal{U}(0.5, 3)$, from which the bounds are derived as
$[\mu - \sigma\sqrt{3},\; \mu + \sigma\sqrt{3}]$. For the \emph{Weibull},
location $\ell \sim \mathcal{U}(0, 2)$, shape $\alpha \sim \mathcal{U}(1, 3)$,
and scale $\lambda \sim \mathcal{U}(0.5, 2)$. For the \emph{Cauchy}, location
$\alpha \sim \mathcal{U}(-5, 5)$ and scale $\beta \sim \mathcal{U}(0.5, 2)$.
For the \emph{Laplace}, location $\mu \sim \mathcal{U}(-5, 5)$ and diversity
$b \sim \mathcal{U}(0.5, 2)$. Each dataset contains $n \in [500, 1500]$ points.

\paragraph{Hard split.}
Two distribution families are drawn independently with replacement from the
same eight families. Each component draws its own sample size $n_k \in [250,
750]$ independently, and parameters for each component are sampled using the
same per-family ranges described above. The two components are concatenated to
form the final dataset of total size $n = n_1 + n_2$.

\paragraph{Astro split.}
Each ground-truth model is one of five initial mass function (IMF) families:
Salpeter, Kroupa, Chabrier, Free-form (wide), and Free-form (tight), with 20
independent realisations generated per family. The number of stellar masses
drawn per realisation is sampled uniformly from $[600, 1500]$, and all
families share $m_{\max} = 120\,M_\odot$. The Salpeter IMF is a
single power law $\xi(m) \propto m^{-\alpha}$ with $\alpha \sim \mathcal{U}(1.9,
2.7)$ on $[0.08, 120]\,M_\odot$. The Kroupa IMF is a three-segment broken
power law with slopes $p_1 \sim \mathcal{U}(0.1, 0.5)$, $p_2 \sim
\mathcal{U}(1.0, 1.6)$, $p_3 \sim \mathcal{U}(1.9, 2.7)$ and fixed break
masses $m_1 = 0.08\,M_\odot$, $m_2 = 0.5\,M_\odot$ on $[0.009,
120]\,M_\odot$; continuity constants are enforced at each break. The Chabrier
IMF joins a log-normal component ($m_{\text{char}} \sim \mathcal{U}(0.15,
0.35)$, $\sigma \sim \mathcal{U}(0.45, 0.70)$) below $1\,M_\odot$ to a power
law ($\alpha \sim \mathcal{U}(1.9, 2.7)$) above $1\,M_\odot$, matched
continuously at the junction. The Free-form IMF is an 8-parameter broken power
law with four slopes and three interior break masses; the wide variant uses
broad exploratory ranges ($a_0 \!\sim\! \mathcal{U}(-0.5, 1.0)$,
$a_1 \!\sim\! \mathcal{U}(0.5, 2.0)$, $a_2 \!\sim\! \mathcal{U}(1.5, 3.0)$,
$a_3 \!\sim\! \mathcal{U}(1.8, 3.5)$; breaks $b_1 \!\sim\! \mathcal{U}(0.06,
0.12)$, $b_2 \!\sim\! \mathcal{U}(0.30, 0.80)$, $b_3 \!\sim\! \mathcal{U}(0.80,
2.00)$; $m_{\min} \!\sim\! \mathcal{U}(0.007, 0.040)$), while the tight variant
uses ranges anchored near canonical Kroupa values ($a_0 \!\sim\!
\mathcal{U}(0.1, 0.5)$, $a_1 \!\sim\! \mathcal{U}(1.0, 1.6)$,
$a_2, a_3 \!\sim\! \mathcal{U}(1.9, 2.7)$; breaks $b_1 \!\sim\!
\mathcal{U}(0.07, 0.09)$, $b_2 \!\sim\! \mathcal{U}(0.40, 0.60)$,
$b_3 \!\sim\! \mathcal{U}(0.80, 1.20)$; $m_{\min} \!\sim\! \mathcal{U}(0.007,
0.012)$). Break masses are sorted to enforce $b_1 < b_2 < b_3$, and
continuity constants are applied at each break.

\subsection{Time Series}

For each $D_{\text{True}_i}$ in the time-series domain, ground-truth signal
parameters are drawn independently and uniformly from predefined ranges, with
the specific ranges and signal family depending on the difficulty split.

\paragraph{Easy split.}
Each ground-truth signal follows a sinusoidal trend model,
\begin{equation}
    y(t) = s \cdot \tilde{t} + A\sin(2\pi f \tilde{t}) + \varepsilon(t),
\end{equation}
where $\tilde{t} \in [0,1]$ is normalized time. Parameters are sampled as
slope $s \sim \mathcal{U}(-2.5,\, 2.5)$, amplitude $A \sim \mathcal{U}(0.2,\,
2.0)$, frequency $f \sim \mathcal{U}(1,\, 12)$ cycles per window, and noise
fraction $\nicefrac{\sigma}{A} \sim \mathcal{U}(0.05,\, 0.15)$ with
$\varepsilon(t) \sim \mathcal{N}(0, \sigma^2)$. Approximately 16\% of series
are further constrained to $s = 0$ to ensure flat-trend instances are
represented. All series are mean-centred after generation.

\paragraph{Hard split.}
A two-level sampling scheme is used. First, a signal category is
drawn---\textit{Complex} with probability $0.4$ or \textit{Rhythmic} with
probability $0.6$. Within \textit{Complex}, a type is drawn uniformly from:
ARMA (an AR(1) random walk), Heteroscedastic (an expanding-variance process
with additive seasonal component), and Logistic (an S-curve trend). Within
\textit{Rhythmic}, a type is drawn uniformly from: synthetic ECG (a Gaussian
P-QRS-T complex template) and Square wave, with period sampled from $\{40, 60,
80\}$ time steps; small Gaussian noise $\varepsilon(t) \sim \mathcal{N}(0,
\sigma^2)$ is added to all rhythmic signals. Pure sine-wave instances are
generated but excluded from the final dataset to prevent overlap with the easy
split.

\paragraph{Astro split (gravitational wave chirps).}
Each ground-truth signal follows a linearly swept-frequency (chirp) model,
\begin{equation}
    y(t) = A(t)\sin\!\left(2\pi\!\left(f_0\, t +
    \tfrac{1}{2}\beta\, t^2\right)\right) + \varepsilon(t),
\end{equation}
giving instantaneous frequency $f(t) = f_0 + \beta t$. Parameters are sampled
as $f_0 \sim \mathcal{U}(0.3,\, 2.5)$, $\beta \sim \mathcal{U}(0.20,\, 1.00)$,
$A \sim \mathcal{U}(0.7,\, 1.8)$, and $\sigma \sim \mathcal{U}(0.07,\, 0.15)$.
For 40 of the 50 series, amplitude is constant, $A(t) = A$; for the remaining
10, an exponential decay envelope $A(t) = A\,e^{-t/\tau}$ with $\tau \sim
\mathcal{U}(2.0,\, 5.0)$ is applied to model inspiral ringdown behaviour.

All time series consist of $N = 600$ equally spaced observations on a daily grid.

\section{\method{} Tool Generation Variants}
 To evaluate whether tools can be effectively accumulated through sequential interaction, we sampled 100 instances from the \protect\Hard{} distribution fitting split, partitioning them equally into 50 train and 50 test examples. We evaluate three conditions with Claude Sonnet 4.6. (1) Accumulate-Only: During a train phase, the model processes training examples sequentially with tool accumulation enabled, initializing each run with the tools generated in all prior runs and selecting either an existing tool or generating a new one. After 50 sequential runs, this process converges to just 4 unique tools. During the test phase, the model is restricted to selecting from these accumulated tools with no new generation permitted and no interaction between runs. (2) Generate-Only: Identical to Accumulate-Only, but the test phase additionally permits the model to generate new tools, which are not shared across test runs. (3) Expert Toolkit + Generation: No train phase is used. Instead, the model is initialized with the default expert toolkit on all 50 test examples while also retaining the ability to generate new tools, again without cross-run sharing. This design isolates three factors: the value of accumulated tools alone, the benefit of combining accumulated tools with on-the-fly generation, and the contribution of generation when paired with a fixed expert toolkit.

\begin{wrapfigure}{r}{0.45\textwidth}
    \centering
    \includegraphics[width=0.43\textwidth]{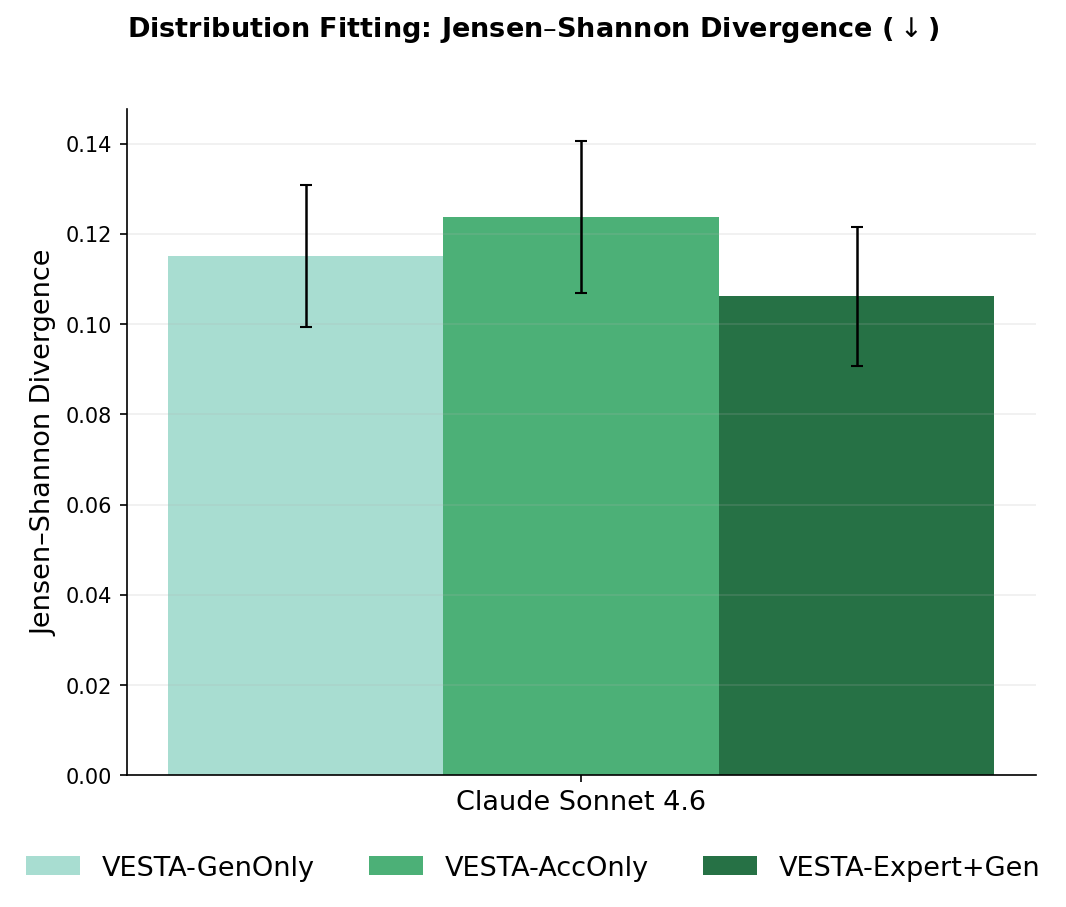}
    \caption{Jensen--Shannon divergence on the \protect\Hard{} distribution fitting split (lower is better) comparing three accumulated toolkit conditions with Claude Sonnet 4.6.}
    \label{fig:accumulated_variant_comparison}
\end{wrapfigure}

Figure~\ref{fig:accumulated_variant_comparison} presents JS divergence scores across the three toolkit conditions on the Hard distribution fitting split. Overall, performance differences across conditions are modest, with all three variants achieving mean JS divergence between 0.106 and 0.124, and overlapping standard errors indicating no statistically significant separation. Nevertheless, a consistent ordering emerges: Expert+Generate achieves the lowest divergence (0.106), followed by Generate-Only (0.115), with Accumulate-Only performing worst (0.124). Two conclusions follow from this pattern. First, tool accumulation alone provides no measurable benefit. Restricting the model to the four auto-accumulated tools at test time yields the weakest performance, suggesting that sequential accumulation over 50 training examples converges to a toolkit that is too narrow to generalize across the full diversity of dataset. Second, on-the-fly tool generation is the primary performance driver. Both conditions that permit test-time generation outperform Accumulate-Only, regardless of whether the base toolkit is accumulated or expert-designed. The marginal advantage of Expert+Generate over Generate-Only further suggests that a well-curated expert toolkit provides a stronger generative prior than one derived through automatic accumulation. Taken together, these results indicate that the current accumulation strategy does not yet yield tools competitive with expert-designed ones, and that allowing the model to generate new tools at test time remains essential for robust distribution fitting performance.

\section{Additional Results on \method{}}
\label{app:additional_results}

\subsection{Statistical Significance of the Results}
\label{app:significance}

We pair-wise test, for each combination of LLM, dataset split, baseline, and \method{} 
variant, whether \method{} significantly outperforms the baseline. Both evaluation 
metrics are treated as independent statistical questions, as: they probe different tasks 
(distribution fitting vs.\ time-series modelling), have different data structures, and warrant different tests. Multiple-comparison corrections are applied within each metric only. We compare to BoxLM and PyVision, our primary baseline approaches.

\paragraph{Distribution fitting (JS divergence).} Per-problem JS values are available 
for every method on the same set of problems, so we use a one-sided paired Wilcoxon 
signed-rank test across problems with $H_1\!:\mathrm{JS}_{\method} < \mathrm{JS}_{\text{baseline}}$.

\paragraph{Time-series modeling (ELPD LOO).} ArviZ's \texttt{loo} returns a point 
estimate $\widehat{\mathrm{ELPD}}$ and its standard error $\widehat{\mathrm{SE}}$ for each 
fit. We use a one-sided $z$-test on 
$\Delta\mathrm{ELPD} \!=\! \mathrm{ELPD}_{\method} \!-\! \mathrm{ELPD}_{\text{baseline}}$ with 
$\mathrm{SE}_\Delta = \sqrt{\widehat{\mathrm{SE}}_V^{\,2} + \widehat{\mathrm{SE}}_B^{\,2}}$. 
This $\mathrm{SE}_\Delta$ is \emph{conservative}: \texttt{loo\_compare} computes the SE 
from per-observation ELPD differences, which is strictly smaller whenever the two ELPDs 
are positively correlated. Our reported 
$p$-values are therefore upper bounds on the true $p$-values.

\paragraph{Multiple comparisons.} Within each metric we conduct $3 \text{ splits} 
\times 3 \text{ LLMs} \times 3 \text{ \method{} variants} \times 2 \text{ baselines} 
= 54$ tests and apply Holm--Bonferroni correction. We use $\alpha = 0.05$ throughout.

\subsection{Summary: aggregate win rates}

Table~\ref{tab:sig_summary} reports, per (variant, baseline, split), how many of the 
3 LLMs yield a statistically significant win for \method{} after Holm correction. 
\method{}'s advantage on time-series modelling is robust: every variant beats PyVision 
in every (LLM, split) cell, and beats Box-LM in 13/18 cells. For distribution fitting, 
significant wins are concentrated on the \Hard{} split, where the absolute JS gap 
between \method{} and the baselines is largest.

\begin{table}[ht]
    \centering
    \footnotesize
    \setlength{\tabcolsep}{4pt}
    \caption{Number of LLMs (out of 3) for which each \method{} variant significantly 
    outperforms each baseline at $\alpha\!=\!0.05$ after Holm correction. Tests are 
    one-sided in the direction ``\method{} beats baseline'' (paired Wilcoxon for JS 
    divergence; $z$-test on $\Delta$ELPD for ELPD LOO). Bold marks cells with 
    $\geq\!1$ significant win.}
    \label{tab:sig_summary}
    \begin{tabular}{l cc cc cc}
      \hline\hline
       & \multicolumn{2}{c}{\textbf{\protect\Easy{}}}
       & \multicolumn{2}{c}{\textbf{\protect\Hard{}}}
       & \multicolumn{2}{c}{\textbf{\protect\Astro{}}} \\
       \cmidrule(lr){2-3}\cmidrule(lr){4-5}\cmidrule(lr){6-7}
       & PyVis. & Box-LM & PyVis. & Box-LM & PyVis. & Box-LM \\
      \hline
      \multicolumn{7}{l}{\textit{Distribution fitting --- JS divergence}} \\
      \method{}-None    & 1/3 & 0/3 & \textbf{2/3} & 0/3          & 0/3          & 0/3 \\
      \method{}-Dynamic & 1/3 & 0/3 & \textbf{2/3} & 2/3          & 1/3          & 0/3 \\
      \method{}-Expert  & 1/3 & 1/3 & \textbf{2/3} & \textbf{1/3} & \textbf{2/3} & \textbf{1/3} \\
      \hline
      \multicolumn{7}{l}{\textit{Time-series modelling --- ELPD LOO}} \\
      \method{}-None    & \textbf{3/3} & \textbf{2/3} & \textbf{2/3} & 0/3          & \textbf{3/3} & \textbf{3/3} \\
      \method{}-Dynamic & \textbf{3/3} & \textbf{2/3} & \textbf{3/3} & \textbf{2/3} & \textbf{3/3} & \textbf{3/3} \\
      \method{}-Expert  & \textbf{3/3} & \textbf{2/3} & \textbf{3/3} & \textbf{2/3} & \textbf{3/3} & \textbf{3/3} \\
      \hline\hline
    \end{tabular}
\end{table}

\subsection{Detailed per-LLM results}

Tables~\ref{tab:sig_js_detail} and~\ref{tab:sig_elpd_detail} give the median JS gap 
and $\Delta$ELPD respectively for every individual comparison, along with stars 
denoting Holm-corrected significance levels.

\begin{table}[ht]
    \centering
    \footnotesize
    \setlength{\tabcolsep}{4pt}
    \caption{Distribution fitting: median JS gap 
    $\Delta\!=\!\mathrm{med}(\mathrm{JS}_{\method}) - \mathrm{med}(\mathrm{JS}_{\text{baseline}})$. 
    Negative values favour \method{}. Stars are Holm-corrected $p$-values from the 
    one-sided paired Wilcoxon test: \textsuperscript{*}\,$p\!<\!0.05$, 
    \textsuperscript{**}\,$p\!<\!0.01$, \textsuperscript{***}\,$p\!<\!0.001$. 
    Significant wins ($p_{\text{Holm}}\!<\!0.05$) in bold.}
    \label{tab:sig_js_detail}
    \begin{tabular}{ll cc cc cc}
      \hline\hline
       & & \multicolumn{2}{c}{\textbf{\protect\Easy{}}}
       & \multicolumn{2}{c}{\textbf{\protect\Hard{}}}
       & \multicolumn{2}{c}{\textbf{\protect\Astro{}}} \\
       \cmidrule(lr){3-4}\cmidrule(lr){5-6}\cmidrule(lr){7-8}
      \textbf{LLM} & \textbf{Method} & PyVis. & Box-LM & PyVis. & Box-LM & PyVis. & Box-LM \\
      \hline
      \multirow{3}{*}{GPT-5.4-mini}
       & \method{}-None    & $+0.084$ & $+0.031$ & $\boldsymbol{-0.144}^{**}$  & $-0.102$                    & $-0.001$ & $-0.009$ \\
       & \method{}-Dynamic & $+0.026$ & $-0.027$ & $\boldsymbol{-0.105}^{*}$   & $-0.063$                    & $-0.023$ & $-0.031$ \\
       & \method{}-Expert  & $-0.047$ & $-0.100$ & $\boldsymbol{-0.148}^{***}$ & $\boldsymbol{-0.106}^{**}$  & $-0.078$ & $-0.086$ \\
      \hline
      \multirow{3}{*}{Claude Sonnet 4.6}
       & \method{}-None    & $-0.024$ & $-0.114$ & $\boldsymbol{-0.137}^{**}$  & $-0.070$ & $-0.034$                  & $-0.022$ \\
       & \method{}-Dynamic & $-0.076$ & $-0.166$ & $\boldsymbol{-0.149}^{***}$ & $-0.082$ & $-0.069$                  & $-0.057$ \\
       & \method{}-Expert  & $-0.076$ & $-0.166$ & $\boldsymbol{-0.136}^{***}$ & $-0.069$ & $\boldsymbol{-0.120}^{*}$ & $-0.108$ \\
      \hline
      \multirow{3}{*}{Kimi K2.5}
       & \method{}-None    & $+0.031$ & $+0.005$ & $-0.030$ & $-0.005$ & $+0.144$ & $+0.279$ \\
       & \method{}-Dynamic & $-0.010$ & $-0.036$ & $-0.083$ & $-0.058$ & $-0.370$ & $-0.235$ \\
       & \method{}-Expert  & $-0.004$ & $-0.030$ & $-0.077$ & $-0.052$ & $-0.126$ & $+0.009$ \\
      \hline\hline
    \end{tabular}
\end{table}

\begin{table}[ht]
    \centering
    \footnotesize
    \setlength{\tabcolsep}{4pt}
    \caption{Time-series modelling: 
    $\Delta\mathrm{ELPD}\!=\!\mathrm{ELPD}_{\method}\!-\!\mathrm{ELPD}_{\text{baseline}}$, 
    in nats (positive favours \method{}). Stars are Holm-corrected $p$-values from 
    the one-sided $z$-test: \textsuperscript{*}\,$p\!<\!0.05$, 
    \textsuperscript{**}\,$p\!<\!0.01$, \textsuperscript{***}\,$p\!<\!0.001$. 
    Significant wins in bold.}
    \label{tab:sig_elpd_detail}
    \begin{tabular}{ll cc cc cc}
      \hline\hline
       & & \multicolumn{2}{c}{\textbf{\protect\Easy{}}}
       & \multicolumn{2}{c}{\textbf{\protect\Hard{}}}
       & \multicolumn{2}{c}{\textbf{\protect\Astro{}}} \\
       \cmidrule(lr){3-4}\cmidrule(lr){5-6}\cmidrule(lr){7-8}
      \textbf{LLM} & \textbf{Method} & PyVis. & Box-LM & PyVis. & Box-LM & PyVis. & Box-LM \\
      \hline
      \multirow{3}{*}{GPT-5.4-mini}
       & \method{}-None    & $\boldsymbol{137}^{***}$ & $\boldsymbol{64.3}^{**}$  & $30$                     & $-39$                     & $\boldsymbol{143}^{***}$ & $\boldsymbol{115}^{***}$ \\
       & \method{}-Dynamic & $\boldsymbol{138}^{***}$ & $\boldsymbol{65.0}^{***}$ & $\boldsymbol{114}^{***}$ & $\boldsymbol{44.8}^{*}$   & $\boldsymbol{161}^{***}$ & $\boldsymbol{133}^{***}$ \\
       & \method{}-Expert  & $\boldsymbol{166}^{***}$ & $\boldsymbol{92.9}^{***}$ & $\boldsymbol{146}^{***}$ & $\boldsymbol{76.5}^{***}$ & $\boldsymbol{164}^{***}$ & $\boldsymbol{136}^{***}$ \\
      \hline
      \multirow{3}{*}{Claude Sonnet 4.6}
       & \method{}-None    & $\boldsymbol{92.7}^{***}$ & $591$ & $\boldsymbol{125}^{***}$ & $398$ & $\boldsymbol{163}^{***}$ & $\boldsymbol{97.7}^{***}$ \\
       & \method{}-Dynamic & $\boldsymbol{118}^{***}$  & $617$ & $\boldsymbol{251}^{***}$ & $524$ & $\boldsymbol{176}^{***}$ & $\boldsymbol{111}^{***}$ \\
       & \method{}-Expert  & $\boldsymbol{145}^{***}$  & $644$ & $\boldsymbol{264}^{***}$ & $537$ & $\boldsymbol{195}^{***}$ & $\boldsymbol{130}^{***}$ \\
      \hline
      \multirow{3}{*}{Kimi K2.5}
       & \method{}-None    & $\boldsymbol{182}^{***}$ & $\boldsymbol{142}^{***}$ & $\boldsymbol{87.2}^{***}$ & $14.8$                   & $\boldsymbol{173}^{***}$ & $\boldsymbol{128}^{***}$ \\
       & \method{}-Dynamic & $\boldsymbol{200}^{***}$ & $\boldsymbol{160}^{***}$ & $\boldsymbol{195}^{***}$  & $\boldsymbol{123}^{**}$  & $\boldsymbol{171}^{***}$ & $\boldsymbol{126}^{***}$ \\
       & \method{}-Expert  & $\boldsymbol{209}^{***}$ & $\boldsymbol{170}^{***}$ & $\boldsymbol{219}^{***}$  & $\boldsymbol{147}^{***}$ & $\boldsymbol{192}^{***}$ & $\boldsymbol{147}^{***}$ \\
      \hline\hline
    \end{tabular}
\end{table}

\subsection{Discussion}

\paragraph{Time-series modelling.} \method{}'s advantage on ELPD LOO is essentially 
universal: 45 of 54 comparisons are significant at $\alpha\!=\!0.05$ after Holm 
correction, with the great majority at $p\!<\!0.001$. The five non-significant 
comparisons against Box-LM on Claude (Easy and Hard splits) are an artefact of 
Box-LM's catastrophically diffuse posterior on Claude --- the $\Delta\mathrm{ELPD}$ 
point estimates are massive ($+398$ to $+644$ nats in favour of \method{}), but the 
standard error on the Box-LM side ($\sim\!360$ nats) is large enough to keep these 
just above the Holm-corrected threshold. Substantively, \method{} clearly wins here; 
the test is simply unable to certify it given how unstable Box-LM's posterior is.

\paragraph{Distribution fitting.} Significant wins on JS divergence are concentrated 
in the \Hard{} split, where the effect size is largest. On \Easy{} and \Astro{}, 
point estimates favour \method{} in nearly every cell (Table~\ref{tab:sig_js_detail}), 
but per-problem variance is high relative to the typical gap, and the Wilcoxon test 
cannot certify the difference at $\alpha\!=\!0.05$ given 54 corrections. The Kimi/\Astro{} 
results are particularly noisy: \method{}-Dynamic posts a $-0.370$ median JS gap vs.\ 
PyVision yet $p_{\text{Holm}}\!=\!1$, because the paired sample is small and skewed. 
We therefore present the JS divergence results as suggestive rather than confirmatory 
on \Easy{} and \Astro{}, and as confirmatory on \Hard{}.

\begin{table}[ht]
    \centering
    \scriptsize
    \setlength{\tabcolsep}{3.5pt}
    \caption{Distribution fitting: posterior quality metrics by LLM, method, and dataset. ELPD LOO ($\uparrow$): posterior predictive log-density on held-out data. Bhatt ($\downarrow$): Bhattacharyya distance between predicted and true distributions. $p_{\text{loo}}$ ($\downarrow$): PSIS-LOO effective number of parameters (lower indicates cleaner posterior geometry). Bold marks the best value within each (LLM, dataset, metric) column.}
    \label{tab:dist_fitting_all}
    \begin{tabular}{ll rrr rrr rrr}
      \hline\hline
      & & \multicolumn{3}{c}{\textbf{\protect\Easy{}}}
        & \multicolumn{3}{c}{\textbf{\protect\Astro{}}}
        & \multicolumn{3}{c}{\textbf{\protect\Hard{}}} \\
      \cmidrule(lr){3-5}\cmidrule(lr){6-8}\cmidrule(lr){9-11}
      \textbf{LLM} & \textbf{Configuration}
        & ELPD & Bhatt & $p_{\text{loo}}$
        & ELPD & Bhatt & $p_{\text{loo}}$
        & ELPD & Bhatt & $p_{\text{loo}}$ \\
      \hline
      \multirow{5}{*}{GPT 5.4-mini}
        & PyVision         & $-1981.48$           & $0.68$          & $4.41$          & $181.31$           & $0.54$          & $6.58$          & $-2426.22$          & $0.40$          & $17.11$         \\
        & Box-LM           & $\mathbf{-1788.34}$ & $0.51$          & $4.68$          & $\mathbf{281.73}$ & $0.41$          & $6.23$          & $-2381.70$          & $0.30$          & $12.23$         \\
        & \method{}-None    & $-1964.96$           & $0.44$          & $5.52$          & $267.33$           & $0.33$          & $5.54$          & $-2359.47$          & $0.21$          & $\mathbf{6.94}$ \\
        & \method{}-Dynamic & $-1860.53$           & $0.43$          & $3.17$          & $279.21$           & $0.36$          & $5.77$          & $-2365.76$          & $0.18$          & $7.19$          \\
        & \method{}-Expert  & $-1853.66$           & $\mathbf{0.41}$ & $\mathbf{3.08}$ & $281.12$           & $\mathbf{0.28}$ & $\mathbf{5.33}$ & $\mathbf{-2344.45}$ & $\mathbf{0.17}$ & $7.17$          \\
      \hline
      \multirow{5}{*}{Claude Sonnet 4.6}
        & PyVision         & $-1997.60$           & $0.49$          & $2.32$          & $249.36$           & $0.57$          & $2.61$          & $-2661.05$           & $0.44$          & $23.36$         \\
        & Box-LM           & $\mathbf{-1707.01}$ & $0.54$          & $3.35$          & $199.42$           & $0.44$          & $6.67$          & $\mathbf{-2288.47}$ & $0.27$          & $11.79$         \\
        & \method{}-None    & $-1856.27$           & $0.51$          & $\mathbf{1.92}$ & $\mathbf{318.42}$ & $0.57$          & $\mathbf{2.59}$ & $-2330.62$           & $\mathbf{0.17}$ & $\mathbf{7.75}$ \\
        & \method{}-Dynamic & $-1848.46$           & $0.97$          & $4.22$          & $291.38$           & $0.32$          & $3.67$          & $-2348.25$           & $0.18$          & $7.82$          \\
        & \method{}-Expert  & $-1883.78$           & $\mathbf{0.44}$ & $2.82$          & $286.56$           & $\mathbf{0.31}$ & $5.05$          & $-2340.52$           & $0.18$          & $13.85$         \\
      \hline
      \multirow{5}{*}{Kimi K2.5}
        & PyVision         & $-2133.54$           & $0.54$          & $16.10$         & $180.66$           & $0.54$          & $8.59$          & $-2458.90$          & $0.34$          & $20.45$         \\
        & Box-LM           & $\mathbf{-1788.34}$ & $0.51$          & $4.68$          & $281.73$           & $0.41$          & $6.23$          & $-2381.70$          & $0.30$          & $12.23$         \\
        & \method{}-None    & $-1840.80$           & $1.13$          & $5.01$          & $\mathbf{417.58}$ & $0.97$          & $\mathbf{1.32}$ & $\mathbf{-2335.40}$ & $0.24$          & $10.42$         \\
        & \method{}-Dynamic & $-1891.31$           & $\mathbf{0.50}$ & $5.19$          & $299.67$           & $0.34$          & $6.91$          & $-2351.49$          & $0.20$          & $\mathbf{8.88}$ \\
        & \method{}-Expert  & $-1839.91$           & $0.70$          & $\mathbf{4.14}$ & $291.38$           & $\mathbf{0.23}$ & $6.57$         & $-2349.39$          & $\mathbf{0.17}$ & $13.76$         \\
      \hline\hline
    \end{tabular}
\end{table}

\subsection{Alternative Evaluation Metrics}
\label{app:alterative_evals}
\subsubsection{Beyond ELPD for Time-Series}
\label{app:subsub-beyond-elpd}
\begin{table}[t]
\centering
\caption{Mean performance comparison across all toolkit configurations and datasets for time series. Metrics: ELPD LOO (higher $\uparrow$), $R^2$ (higher $\uparrow$), CRPS (lower $\downarrow$); We use \textsc{PyVision} as our baseline.}
\label{tab:comprehensive_comparison}
\vspace{0.5em}
\resizebox{\textwidth}{!}{%
\begin{tabular}{ll|rrr|rrr|rrr}
\toprule
 & & \multicolumn{3}{c|}{\textbf{Easy}} & \multicolumn{3}{c|}{\textbf{Astro (Chirp)}} & \multicolumn{3}{c}{\textbf{Hard}} \\[0.25em]
\textbf{LLM} & \textbf{Configuration} & ELPD & $R^2$ & CRPS & ELPD & $R^2$ & CRPS & ELPD & $R^2$ & CRPS \\
\midrule
\multirow{5}{*}{Claude Sonnet 4.6} & \textsc{Vesta} No-Toolkit & 30.2 & 0.870 & 0.1127 & 10.1 & 0.675 & 0.1949 & -210.7 & 0.819 & 0.6891 \\
 & \textsc{Vesta} Dynamic-Toolkit & 55.9 & 0.965 & 0.0749 & 23.3 & 0.842 & 0.1085 & -85.1 & 0.920 & 0.3537 \\
 & \textsc{Vesta} Expert-Toolkit & 82.7 & 0.981 & 0.0660 & 42.3 & 0.909 & 0.0798 & -71.8 & 0.938 & 0.3689 \\
 & \textsc{PyVision} & -62.5 & $\approx 0$ & 0.490 & -153.1 & $\approx 0$ & 0.5052 & -335.9 & $\approx 0$ & 4.833 \\
 & \textsc{BoxLM} & \textbf{\underline{-561.0}} & 0.639 & 0.235 & -87.6 & 0.342 & 0.3318 & \textbf{\underline{-608.6}} & 0.573 & 1.1383 \\
\midrule
\multirow{5}{*}{GPT 5.4 mini} & \textsc{Vesta} No-Toolkit & 33.7 & 0.769 & 0.1864 & 26.2 & 0.791 & 0.1238 & -233.4 & 0.830 & 0.6218 \\
 & \textsc{Vesta} Dynamic-Toolkit & 34.4 & 0.812 & 0.1244 & 44.5 & 0.882 & 0.0875 & -149.4 & 0.844 & 0.4746 \\
 & \textsc{Vesta} Expert-Toolkit & 62.3 & 0.962 & 0.0702 & 47.5 & 0.891 & 0.0811 & -117.7 & 0.948 & 0.4445 \\
 & \textsc{PyVision} & -103.4 & 0.085 & 0.512 & -116.5 & 0.031 & 0.448 & -263.4 & $\approx 0$ & 4.4379 \\
 & \textsc{BoxLM} & -30.6 & 0.434 & 0.3438 & -88.6 & 0.024 & 0.4523 & -194.2 & 0.412 & 1.5097 \\
\midrule
\multirow{5}{*}{Kimi K2.5} & \textsc{Vesta} No-Toolkit & 80.7 & 0.830 & 0.1481 & 36.6 & 0.857 & 0.1061 & -222.3 & 0.741 & 0.8551 \\
 & \textsc{Vesta} Dynamic-Toolkit & 98.7 & 0.909 & 0.0936 & 34.6 & 0.884 & 0.1001 & -114.1 & 0.907 & 0.3651 \\
 & \textsc{Vesta} Expert-Toolkit & 108.1 & 0.962 & 0.0720 & 55.5 & 0.965 & 0.0559 & -90.1 & 0.969 & 0.2776 \\
 & \textsc{PyVision} & -101.1 & $\approx 0$ & 0.512 & -136.7 & $\approx 0$ & 0.4899 & -309.5 & $\approx 0$ & 4.0653 \\
 & \textsc{BoxLM} & -61.6 & 0.462 & 0.3249 & -91.1 & 0.004 & 0.4575 & -237.1 & 0.401 & 1.7423 \\
\bottomrule
\end{tabular}%
}
\end{table}

\noindent Estimating the Expected Log Pointwise Predictive Density (ELPD) for Gaussian Process based Time Series models 
in \texttt{PyMC} is computationally demanding because it necessitates full MCMC sampling. Unlike simpler parametric models, GPs require the inversion of 
an $n \times n$ covariance matrix - a process with $O(n^3)$ complexity - at every gradient 
evaluation within the NUTS sampler. To mitigate this cubic scaling bottleneck, we employ fixed-space sub-sampling. By selecting a 
representative subset of the time series data at regular intervals, we reduce the effective 
$n$ and this significantly speeds up the evaluation. For consistency, we measure the $R^{2}$, CRPS are calculated in the same manner.

For the baseline evaluations, BoxLM generally performs better than PyVision, with the exception of a small number of anomalous cases in Claude-generated PyMC code. Specifically, a few extremely large ELPD values substantially inflate the mean performance metrics. Excluding these 7 outliers (out of 630 total samples) reduces the mean ELPD for Claude Sonnet 4.6 + BoxLM to \textbf{40.4} on the Easy dataset and \textbf{-150.6} on the Hard dataset, bringing performance closer for Easy, and ahead for Hard dataset to the corresponding No-Toolkit configuration. \par

In contrast, PyVision exhibits substantially higher failure rates during model fitting. Across datasets, approximately 40--60\% of generated samples fail either due to code execution issues or unsuccessful MCMC fitting, representing the highest failure rate among all evaluated approaches. In the remaining successful runs, the fitted models frequently collapse to near mean-value predictions, resulting in poor overall ELPD and near-zero $R^{2}$ values. These failures are largely attributable to PyVision's implementation strategy, which is primarily optimized around MAP-based fitting and does not consistently translate to stable MCMC-based inference.

\subsubsection{Beyond JS-Divergence for IMFs}
\label{app:subsub-beyond-js}
For IMF fitting, astronomers often care more whether the recovered power-law slopes and break masses are correct as opposed to the exact distribution function. To stay consistent with the other two distribution fitting tasks, we evaluated the Astro split with Jensen-Shannon divergence in the main results. Here we add an evaluation metric for the Astro split by inferring slope parameters from the predicted PyMC models. We report each VESTA configuration with the Claude Sonnet 4.6 backbone.

The central parameter we study is the logarithmic slope
\begin{equation}
    \Gamma(m) = -\frac{d \log f(m)}{d \log m},
\end{equation}
which equals $\alpha$ exactly on a power-law segment, but remains well-defined for any fitted family (including fits with no power-law component). From $\Gamma$ we derive an effective slope parameter $\alpha_{\text{eff}}$ using weighted least squares of $\log f$ on $\log m$ over the high-mass window ($m > 1\,M_\odot$), computed identically on the fitted and ground-truth IMFs. We define the \textbf{$\alpha_{\text{eff}}$ Median Absolute Error} metric on the Astro Distribution Fitting task as the  ``Median absolute error in the effective high-mass slope against the ground-truth IMF''.

We find that results are consistent with Jensen-Shannon divergence, but this metric tends to be more favorable for VESTA-Dynamic. VESTA-Dynamic recovers the Salpeter slope over high masses in nearly all cases ($|\Delta \alpha_{\text{eff}}| = 0.030$ on Salpeter instances). VESTA with no toolkit fails badly on the same instances ($|\Delta \alpha_{\text{eff}}| = 6.21$), with only 14\% of fits containing a power-law tail at all. Results for all IMF families are shown in Table~\ref{tab:imf-alpha}.

\begin{table}[h]
    \centering
    \caption{Median absolute error in $\alpha_{\text{eff}}$ over the high-mass window ($m > 1\,M_\odot$). Lower is better; best per row in bold.}
    \label{tab:imf-alpha}
    \begin{tabular}{lccc}
        \toprule
        IMF Family & VESTA-None & VESTA-Expert & VESTA-Dynamic \\
        \midrule
        Salpeter          & 6.213          & 0.899          & \textbf{0.030} \\
        Kroupa            & \textbf{1.171} & 1.345          & 1.343          \\
        Chabrier          & 6.491          & \textbf{1.367} & 4.262          \\
        Freeform (tight)  & 1.472          & \textbf{1.248} & 1.288          \\
        Freeform (wide)   & 1.298          & 0.859          & \textbf{0.805} \\
        \bottomrule
    \end{tabular}
\end{table}

\subsection{Impact of the Iteration Budget in VESTA}
\label{app:iteration_budget}

All VESTA variants are run with a fixed iteration budget of $N=5$. In this section, we examine where the best performing models occur in this fixed budget. We study this to answer whether VESTA-Expert outperforming VESTA-Dynamic could be an artifact of expert tools being simpler to interpret in fewer steps. 

Step 0 is the zero-shot proposal with no refinement, and Step $i$ refers to the $i^{\text{th}}$ step in the VESTA iterative procedure. For each instance, we identify the step with the lowest AIC among VESTA's 5 candidates and the zero-shot baseline, and report the percentage of instances for which the best PyMC model occurred at each step. Results for all three toolkit modes and backbone models are shown in Table~\ref{tab:best_step}.

\begin{table}[h]
\centering
\caption{Percentage of instances for which the best PyMC model (lowest AIC) occurred at each step. Step 0 is the zero-shot proposal; Steps 1-5 are VESTA refinement iterations.}
\label{tab:best_step}
\begin{tabular}{l ccc ccc ccc}
\toprule
 & \multicolumn{3}{c}{VESTA-None} & \multicolumn{3}{c}{VESTA-Expert} & \multicolumn{3}{c}{VESTA-Dynamic} \\
\cmidrule(lr){2-4} \cmidrule(lr){5-7} \cmidrule(lr){8-10}
Step & Claude & GPT & Kimi & Claude & GPT & Kimi & Claude & GPT & Kimi \\
\midrule
0 (0-Shot) & 10 & 3 & 8  & 8  & 2  & 10 & 2  & 4  & 5  \\
1          & 3  & 9 & 1  & 1  & 5  & 1  & 14 & 16 & 17 \\
2          & 1  & 7 & 0  & 2  & 5  & 3  & 7  & 3  & 10 \\
3          & 2  & 4 & 2  & 4  & 2  & 3  & 6  & 3  & 9  \\
4          & 1  & 1 & 2  & 6  & 3  & 1  & 2  & 1  & 11 \\
5          & 83 & 76 & 87 & 79 & 83 & 82 & 69 & 73 & 48 \\
\bottomrule
\end{tabular}
\end{table}

Zero-shot proposals rarely produce the best model.
A zero-shot baseline provides the best model in only 3-10\% of cases, even with the VESTA-None variant where no tools are utilized. With VESTA-Dynamic, the number of cases where the zero-shot proposal improves over VESTA is reduced further to 2-5\%.

If VESTA-Dynamic had a higher concentration of best-performing models at Step 5, that would indicate that the complex generated tools need more iterations to process successfully. We find the opposite. Across all three backbone models, VESTA-Dynamic has the lowest share of best models at the final index. The starkest case is Kimi-2.5, where 82/100 of the best PyMC models occur at Step 5 with VESTA-Expert, but only 48/100 with VESTA-Dynamic. This indicates that VESTA-Expert is potentially more constrained by the $N=5$ budget than VESTA-Dynamic. Further, VESTA-Dynamic tends to converge earlier, with a spike at Step 1 that is absent for VESTA-Expert.

We argue that these observations are sensible given the structure of the toolkits. The generated tools in VESTA-Dynamic are predominantly multi-panel and deliver a large amount of information in a single call. In contrast, Expert tools are designed to perform a single function and must be composed across several iterations. Compared to using no tools, VESTA-Dynamic also converges to the best solution faster. For Kimi-2.5, the number of best models at Step 5 is reduced from 87/100 to 48/100 when the agent is allowed to dynamically generate tools.

This analysis shows that (1) a zero-shot setting rarely produces a better model than VESTA, and (2) equipping VESTA with dynamic tool generation helps agents find optimal solutions at earlier iterations. Consequently, the finding that VESTA-Expert outperforms VESTA-Dynamic in most cases is unlikely to be an artifact of the iteration budget. The gap is also not explained by dynamic tools failing to improve PyMC model predictions, since VESTA-Dynamic rarely has its best model at Step 0. We note that all claims in this paper hold under a fixed and equal iteration budget of $N=5$.

\section{Expert Tools in Detail}
\label{app:tools}
\subsection{Distribution Fitting}
\begin{enumerate}
    \item \textbf{CalculateMoments}: Computes the mean, variance, skewness, and excess 
    kurtosis of the input data. Returns a JSON artifact with a plain-language interpretation 
    to guide distribution selection, including symmetry hints (e.g., right-skewed data 
    suggests Gamma, Lognormal, or Weibull families) and tail-weight hints (e.g., leptokurtic 
    data suggests Student-$t$, Cauchy, or Laplace families).

    \item \textbf{Histogram}: Plots a histogram of the empirical data with the fitted 
    distribution's probability density function (PDF) overlaid. Handles both single 
    distributions and mixtures by summing component PDFs weighted by their mixture weights. 
    Provides an immediate visual check of whether the model captures the overall shape, 
    modality, and spread of the data. When no model has been fitted yet, produces a raw 
    histogram without an overlay.

    \item \textbf{SegmentDistributionsAndCalculateMoments}: Segments the data into a 
    specified number of mixture components using a Gaussian Mixture Model (GMM), then computes 
    per-component moments (mean, variance, skewness, kurtosis). Produces both a segmentation 
    image with a total mixture overlay and a JSON summary of per-component statistics with 
    distribution family hints. Most useful when the histogram appears multimodal.

    \item \textbf{QQPlot}: Generates a Quantile-Quantile (Q-Q) plot comparing empirical 
    data quantiles to theoretical quantiles from the currently fitted distribution. Linearity 
    indicates a good fit; S-shaped curvature signals tail mismatch; one-sided curvature 
    suggests skew; and sharp tail departures may indicate outliers or heavier tails than 
    the model captures.

    \item \textbf{PlotTailsTransform}: Produces log-log and semi-log complementary CDF 
    (CCDF) plots to diagnose tail behavior. A straight line on the log-log plot indicates 
    power-law or Pareto-type heavy tails, while a straight line on the semi-log plot indicates 
    exponential decay. Useful for distinguishing heavy-tailed from light-tailed distributions 
    when the histogram alone is ambiguous.

    \item \textbf{ProbabilityPlot}: Generates a probability plot comparing the empirical 
    CDF to the fitted distribution's theoretical CDF. A consistent horizontal shift indicates 
    a mis-specified location parameter; a slope mismatch indicates a scale misfit; and 
    systematic tail deviations suggest distributional misfit. Also reports a 
    Kolmogorov-Smirnov (KS) statistic for quantitative goodness-of-fit assessment.
\end{enumerate}

\subsection{Time Series}

\begin{table}[t]
\centering
\caption{Per-sample tool coverage: percentage of samples in which each Expert tool is invoked at least once during the iterative refinement loop.}
\label{tab:per-sample-coverage}
\begin{tabular}{l c c c c}
\toprule
\textbf{Tool} & \textbf{Easy} & \textbf{Astro (Chirp)} & \textbf{Hard} & \textbf{Overall} \\
\midrule
\texttt{get\_dominant\_period}          & 100.0\% & 100.0\% & 99.1\% & 99.5\% \\
\texttt{fit\_vs\_actuals\_w/\_resid.}   & 100.0\% & 96.0\%  & 48.2\% & 71.9\% \\
\texttt{residuals\_auto\_corr.\_plot}   & 2.0\%   & 10.0\%  & 54.5\% & 31.4\% \\
\texttt{residuals\_auto\_corr.\_score}  & 0.0\%   & 0.0\%   & 6.4\%  & 3.3\%  \\
\texttt{fit\_vs\_actuals (default at step-0)}               & 100.0\%   & 100.0\%   & 100\%  & 100\%  \\
\bottomrule
\end{tabular}
\end{table}

\begin{enumerate}
    \item \textbf{GetDominantPeriod}: Extracts the dominant period from the time series 
    using Fast Fourier Transform (FFT) analysis. Returns a plain-text summary of the detected 
    period. Most useful when Periodic or PeriodicComplex kernels are under consideration and 
    the period has not yet been numerically determined. The result is available in the 
    subsequent feedback iteration.

    \item \textbf{FitVsActuals}: Produces a visual overlay of the Gaussian Process (GP) 
    fit on the raw time series data. Essential for visually assessing whether the model 
    adequately captures the underlying trend and seasonality while appropriately discounting 
    noise. Falls back to a raw series plot if no model has been fitted yet.

    \item \textbf{FitVsActualsWithResidualsDistribution}: Generates a combined plot 
    showing the GP fit overlaid on the observed time series alongside the distribution of 
    residuals. Used to assess whether residuals resemble white noise; a broadly normal 
    residual distribution is indicative of a well-specified model. Falls back to a raw 
    series plot if no model has been fitted yet.

    \item \textbf{ResidualsAutoCorrelationPlot}: Produces an Autocorrelation Function 
    (ACF) plot of the model residuals to check for temporal independence. Significant spikes 
    above the confidence band indicate that the model is failing to capture some latent 
    structure in the data. Falls back to a raw series plot if no model has been fitted yet.

    \item \textbf{ResidualsAutoCorrelationScore}: Applies the Ljung-Box statistical test 
    to assess residual independence numerically. A $p$-value greater than $0.05$ indicates 
    that the residuals are consistent with white noise, suggesting an adequate model fit. 
    Returns a plain-text summary and requires a fitted model to produce meaningful results.
\end{enumerate}

\subsubsection{Feedback Modality and Adaptive Selection}
The static and dynamic toolkits differ fundamentally in their feedback modality. The expert toolkit provides narrow, targeted signals---such as returning a single scalar for dominant frequency---which allows the VLM to rapidly integrate findings and converge quickly on structurally simple datasets (Easy, Chirp). 

In contrast, the dynamic toolkit generates comprehensive, multi-panel visualizations that synthesize frequency estimation, residual analysis, and harmonic structure into a single artifact. While this rich feedback demands more visual interpretation from the VLM, it is critical for diagnosing complex signals where isolated scalars are insufficient. 

As shown in Table~\ref{tab:per-sample-coverage}, the VLM dynamically adapts its tool selection to the underlying problem complexity. Simple goodness-of-fit overlays dominate the early steps for clean periodic datasets. However, when applied to the Hard dataset, the agent heavily prioritizes autocorrelation and residual diagnostics to detect more subtle, persistent temporal dependencies.

\subsubsection{Comparison with PyVision Baseline}
Apart of choice of modalities, and complexity of serving the tool (via panels), both dynamic-toolkit and expert-toolkit lean into visual feedback. On the other hand, PyVision also generates dynamic tools on the fly, but they diverge significantly in diagnostic focus, output density, and tool lifecycle.

\paragraph{Diagnostic Category Coverage.} 
As detailed in Table~\ref{tab:category-coverage}, both systems heavily utilize Lag/Autocorrelation and Frequency Estimation (${\sim}$85--94\%). However, VESTA's dynamic toolkit exhibits a massive preference for Residual Analysis (91.9\% vs.\ PyVision's 11.9\%) and exclusively employs Statistical Tests (28.6\% vs.\ 0\%). Because residual structure is the primary indicator of GP kernel misspecification, this gap directly impacts model selection quality.

\begin{table}[htbp]
\centering
\caption{Diagnostic category coverage (\% of samples with $\geq$1 tool in category). PV = PyVision.}
\label{tab:category-coverage}
\resizebox{\textwidth}{!}{%
\begin{tabular}{lrrrrrrrr}
\toprule
\multirow{2}{*}{\textbf{Category}} & \multicolumn{2}{c}{\textbf{Easy}} & \multicolumn{2}{c}{\textbf{Astro (Chirp)}} & \multicolumn{2}{c}{\textbf{Hard}} & \multicolumn{2}{c}{\textbf{Overall}} \\
\cmidrule(lr){2-3} \cmidrule(lr){4-5} \cmidrule(lr){6-7} \cmidrule(lr){8-9}
 & VESTA & PV & VESTA & PV & VESTA & PV & VESTA & PV \\
\midrule
Residual Analysis      & 92.0 & 22.0 & 80.0 &  0.0 & 97.3 & 12.7 & \textbf{91.9} & \textbf{11.9} \\
Lag / Autocorrelation  & 98.0 & 98.0 & 92.0 & 94.0 & 84.5 & 84.5 & 89.5 & 90.0 \\
Frequency Estimation   & 100.0 & 100.0 & 100.0 & 100.0 & 71.8 & 89.1 & 85.2 & 94.3 \\
Statistical Tests      & 34.0 &  0.0 & 32.0 &  0.0 & 24.5 &  0.0 & \textbf{28.6} & \textbf{0.0} \\
\bottomrule
\end{tabular}%
}
\end{table}

\section{Analysis of Generated Tools Not in the Expert Toolkit}
\label{app:detailed_tool_analysis}
\subsection{Tool Persistence and Reuse.}
A key architectural advantage of VESTA's dynamic-toolkit is the persistence of its generated functions. Within a single problem instance, 79.5\% of VESTA's dynamically generated tools are re-invoked across subsequent analysis steps (averaging 1.87 calls per tool; Table~\ref{tab:tool-reuse}). This capability is most useful for tools that evaluate model fit, allowing the agent to iteratively reassess model fit using a consistent baseline. Consider the case study shown in Figure~\ref{fig:tool_ex}. As \method{} iteratively refines its distributional hypothesis (from exponential to log-normal), the tool can be re-invoked with an updated PyMC model to regenerate Panel 1's Q-Q plot against the new candidate distribution without incurring additional tool-writing cost. Further, the remaining diagnostic panels act as a consistent reference frame, meaning \method{} does not have to regenerate tools to re-examine the underlying data when evaluating a new distributional hypothesis. PyVision treats generated code as single-use execution cells, entirely lacking a mechanism for revisitation.

\begin{table}[htbp]
\centering
\caption{VESTA tool reuse statistics within single problem instances.}
\label{tab:tool-reuse}
\begin{tabular}{lrrr}
\toprule
\textbf{Dataset} & \textbf{Tools} & \textbf{Reused $\geq$1$\times$} & \textbf{Avg Invocations} \\
\midrule
\protect\Easy{}    &  82 & 64 (78.0\%) & 1.93 \\
\protect\Hard{} & 148 & 126 (85.1\%) & 1.75 \\
\protect\Astro{}   &  73 & 51 (69.9\%) & 2.07 \\
\midrule
\textbf{Overall} & \textbf{303} & \textbf{241 (79.5\%)} & \textbf{1.87} \\
\bottomrule
\end{tabular}
\end{table}

\subsection{Distribution Fitting}
\begin{enumerate}
    \item \textbf{diagnostic\_fit\_checks:} Naming a concrete model family (gaussian, gamma, lognormal, Pareto, Weibull, etc.) and trying it on the data. Each family encodes different assumptions. These tools allow for typically allow for a visual comparison of multiple model families at once. Occasionally, we observe some single use model fitting.  
    
    \item \textbf{information\_criteria:} Numerical scores that rank competing fits while penalizing model complexity. Beyond simply fitting and visualizing models, AIC and BIC provide quantitative fit metrics.  

 \begin{table}[t]
    \centering
    \footnotesize
    \setlength{\tabcolsep}{4pt}
    \caption{Analysis of functions in \method{}-generated tools that are not contained in the expert toolkit for Distribution Fitting.}
    \label{tab:heavytail_tools}
    \begin{tabular}{l rrrr}
      \hline\hline
      \textbf{Function}
        & \textbf{\protect\Easy{}}
        & \textbf{\protect\Hard{}}
        & \textbf{\protect\Astro{}}
        & \textbf{All} \\
      \hline
      Diagnostic Fit Checks  & \textbf{100.0} & 96.9          & \textbf{100.0} & 98.4 \\
      Information Criteria   & \textbf{82.1}  & 73.8          & 65.6           & 73.7 \\
      MLE Fitting            & 69.6           & 36.2          & \textbf{67.2}  & 51.4 \\
      Mean Excess Plot       & 5.4            & 7.7           & \textbf{36.1}  & 14.2 \\
      Hill Estimator         & 0.0            & 1.5           & \textbf{6.6}   & 2.4  \\
      Box--Cox               & \textbf{1.8}   & 0.0           & 0.0            & 0.4  \\
      D'Agostino Normal Test & 0.0            & \textbf{0.8}  & 0.0            & 0.4  \\
      Shapiro--Wilk          & 0.0            & \textbf{0.8}  & 0.0            & 0.4  \\
      \hline\hline
    \end{tabular}
\end{table}

    \item \textbf{mle\_fitting:} Maximum likelihood estimation: choosing the parameter values ($\mu$, $\sigma$, shape, scale, \ldots) that maximize the probability of observing the data under the chosen family. This is the how you actually of fit models, distinct from what models we want to test in \texttt{diagnostic\_fit\_checks}. MLE gives you the canonical ``best'' parameters under a given family, plus a likelihood value that feeds directly into AIC/BIC. 

    \item \textbf{mean\_excess\_plot} Plots the conditional expectation $\mathbb{E}[X - u \mid X > u]$ against threshold $u$. For the Generalized Pareto distribution this function is linear in $u$, so a straight line in the upper tail signals a GPD-like tail and tells you where the ``extreme regime'' begins. This is a tail-diagnostic that complements \textbf{diagnostic\_fit\_checks}. These test help it \method{} decide if the tail behaves as that family predicts and at what threshold the heavy-tail regime kicks in. The 36\% rate on Astro vs.\ $\sim$5\% elsewhere is a sign the model recognizes astronomical data as tail-dominated.

    \item \textbf{hill\_estimator} Estimates the tail index $\alpha$ of a heavy-tailed distribution from the largest $k$ order statistics, giving a concrete number for ``how heavy'' the tail is. A Hill plot ($\hat{\alpha}$ vs.\ $k$) lets you check stability and pick a sensible threshold. This refines a Pareto/power-law fit by pinning down its single most important parameter, and serves as a sanity check on MLE-fit tail parameters --- if MLE on a full-data Pareto fit and a Hill estimate on the upper tail disagree, the body of the distribution is contaminating your tail estimate.

    \item \textbf{shapiro\_wilk} A formal hypothesis test for whether data come from a normal distribution. If Shapiro-Wilk rejects normality strongly, that rules out the normal family in \texttt{diagnostic\_fit\_checks}. 

    \item \textbf{box\_cox} A parametric family of power transforms $y = (x^{\lambda} - 1)/\lambda$ that searches for the $\lambda$ making the transformed data closest to normal and can be useful when working with exotic, heavy-tailed distributions.
\end{enumerate}

\begin{table}[t]
    \centering
    \footnotesize
    \setlength{\tabcolsep}{4pt}
    \caption{Analysis of functions in \method{}-generated tools that are not contained in the expert toolkit for Distribution Fitting.}
    \label{tab:timeseries_tools}
    \begin{tabular}{l rrr}
      \hline\hline
      \textbf{Function}
        & \textbf{\protect\Easy{}}
        & \textbf{\protect\Hard{}}
        & \textbf{\protect\Astro{}} \\
      \hline
      Density Visualization      & 51.9          & \textbf{67.6} & 33.8 \\
      Distribution Shape Summary & 38.3          & \textbf{50.3} & 30.9 \\
      Moving Average             & \textbf{30.9} & 25.5          & 27.9 \\
      Regime Switching Analysis  & 2.5           & \textbf{30.3} & 0.0 \\
      Heavy Tail Diagnostics     & 9.9           & \textbf{20.7} & 5.9 \\
      QQ Plot                    & \textbf{13.6} & 9.7           & 10.3 \\
      Shapiro--Wilk              & \textbf{17.3} & 6.2           & 11.8 \\
      Stationarity Tests         & 1.2           & \textbf{11.7} & 4.4 \\
      Differencing               & 4.9           & \textbf{10.3} & 2.9 \\
      ARIMA                      & 0.0           & \textbf{5.5}  & 0.0 \\
      Change Point Detection     & 0.0           & \textbf{4.1}  & 1.5 \\
      Information Criteria       & 0.0           & \textbf{4.8}  & 0.0 \\
      MLE Fitting                & 0.0           & \textbf{1.4}  & 0.0 \\
      Kolmogorov--Smirnov        & 0.0           & 0.7           & \textbf{1.5} \\
      Log Transform              & 0.0           & 0.7           & \textbf{1.5} \\
      Jarque--Bera               & \textbf{1.2}  & 0.0           & 0.0 \\
      ECDF                       & 0.0           & \textbf{0.7}  & 0.0 \\
      \hline\hline
    \end{tabular}
\end{table}

\subsection{Time Series}
\begin{enumerate}

    \item \textbf{density\_visualization:} Overlays a histogram with a kernel density estimate (KDE) to give a non-parametric picture of the marginal distribution of a time series. This is typically the first diagnostic \method{} reaches for: before committing to any parametric family, the histogram/KDE pair reveals modality, skew, and gross tail behavior at a glance. The KDE bandwidth controls the bias--variance tradeoff in smoothing; narrower bandwidths reveal fine structure while wider ones expose the global shape.

    \item \textbf{distribution\_shape\_summary:} Computes the first four standardized moments (mean, variance, skewness, and excess kurtosis) along with the coefficient of variation $\text{CV} = \sigma/\mu$. These scalars compress the shape of an entire distribution into a small, comparable summary. Skewness distinguishes asymmetric processes (e.g.\ log-normal dwell times) from symmetric ones. Kurtosis flags heavy tails before any formal test is run. This is the numerical complement to \textbf{density\_visualization}.

    \item \textbf{moving\_average:} Computes a rolling mean $\bar{x}_t = \frac{1}{w}\sum_{i=0}^{w-1} x_{t-i}$ over a sliding window of width $w$. This is low-pass filter suppresses high-frequency noise and reveals the underlying trend or long-timescale signal. \method{} uses it both as a standalone smoother and as a preprocessing step before stationarity tests or ARIMA fitting, since many parametric models assume a trend-free series.

    \item \textbf{regime\_switching\_analysis:} A grouped capability covering Gaussian Mixture Models (GMMs), bimodality and multimodality detection, $k$-means clustering on temporal states, Hidden Markov Models (HMMs), and log-normal fits to dwell times. Together these tools test the amount of time a signal spends in qualitatively distinct states?, GMMs and $k$-means partition the amplitude space into clusters; HMMs additionally impose a Markov transition structure on state sequences; dwell-time fits characterize how long the process stays in each regime. This matters for telegraph-noise or switching processes where a single stationary distribution is a category error. The high rate on \protect\Hard{} reflects that harder series are more likely to be genuine multi-state processes.

    \item \textbf{heavy\_tail\_diagnostics:} A grouped capability covering Pareto/power-law identification, Generalized Extreme Value (GEV) fits, Hill estimator calls, and mean-excess plots; see the analogous entries in the distribution-fitting appendix for mathematical detail on the latter two. In the time series context, \method{} applies these when the marginal distribution of the series itself (rather than residuals) appears heavy-tailed, e.g.\ burst-like signals or processes with rare but extreme excursions.

    \item \textbf{qq\_plot:} Plots empirical quantiles $F_n^{-1}(p)$ against theoretical quantiles $F_\theta^{-1}(p)$ for a candidate family $F_\theta$. These are typically used to measure the distribution of residuals.

    \item \textbf{shapiro\_wilk:} A formal hypothesis test of the null that the data are drawn from a normal distribution, with test statistic $W = (\sum a_i x_{(i)})^2 / \sum (x_i - \bar{x})^2$ where $x_{(i)}$ are order statistics and $a_i$ are weights derived from the expected normal order statistics. In the time series context, \method{} typically applies this to residuals after trend/seasonality removal to verify that the generating process is Gaussian, which is a prerequisite for classical ARIMA inference.

    \item \textbf{stationarity\_tests:} A grouped capability covering the Augmented Dickey--Fuller (ADF) test for a unit root and the KPSS test for level/trend stationarity. These are complementary. ADF tests the null of a unit root while KPSS tests the null of stationarity, so using both guards against the failure mode of each test individually. Stationarity is a prerequisite for most parametric time series models; this capability gates whether \method{} proceeds to ARIMA fitting or first applies differencing or detrending.

    \item \textbf{differencing:} Computes the lag-$d$ difference $\nabla^d x_t = x_t - x_{t-d}$, most commonly $d=1$, the first difference. This is the standard cure for a unit root: if the ADF test fails to reject non-stationarity, first-differencing typically removes a stochastic trend and produces a stationary series suitable for ARIMA fitting. Seasonal differencing ($d = s$, the period) removes periodic non-stationarity. \method{} uses this as a preprocessing step that is directly triggered by stationarity test results.

    \item \textbf{arima:} Fits an Autoregressive Integrated Moving Average model $\phi(B)(1-B)^d x_t = \theta(B)\epsilon_t$, where $B$ is the backshift operator, $p$ autoregressive terms capture dependence on recent values, $d$ differences remove trends, and $q$ moving-average terms model the residual autocorrelation structure. ARIMA is the workhorse parametric model for univariate stationary time series; it is typically applied after stationarity tests confirm (or differencing achieves) stationarity, and model order $(p,d,q)$ is selected via AIC/BIC.

    \item \textbf{change\_point\_detection:} Identifies times $\tau$ at which the statistical properties of the series, mean, variance, or autocorrelation structure, shift abruptly. Unlike \textbf{regime\_switching\_analysis}, which models soft probabilistic membership in states, change-point methods partition the timeline into contiguous segments and locate the breakpoints. This is appropriate when the series undergoes irreversible structural shifts (e.g.\ a one-time intervention) rather than recurrent toggling between states.

    \item \textbf{information\_criteria:} Computes AIC $= 2k - 2\hat{\ell}$ and BIC $= k\ln n - 2\hat{\ell}$, where $k$ is the number of free parameters, $n$ the sample size, and $\hat{\ell}$ the maximized log-likelihood. These penalize model complexity to prevent overfitting and give you a principled basis for comparing models of different orders (e.g.\ ARIMA$(1,1,1)$ vs.\ ARIMA$(2,1,2)$) or different parametric families on the same data. AIC favors predictive accuracy while BIC applies a stronger complexity penalty and is consistent for model selection.

    \item \textbf{mle\_fitting:} See the analogous entry in the distribution-fitting appendix. In the time series context, MLE is used both for fitting marginal distributions (e.g.\ the amplitude distribution of a stationary series) and for estimating ARIMA parameters via conditional or exact likelihood. The resulting likelihood value feeds directly into AIC/BIC for model comparison.

    \item \textbf{kolmogorov\_smirnov:} A nonparametric goodness-of-fit test based on the supremum distance $D_n = \sup_x |F_n(x) - F_0(x)|$ between the empirical CDF and a hypothesized distribution $F_0$. Unlike Shapiro--Wilk, the KS test is distribution-free and applies to any fully specified $F_0$, making it useful when \method{} wants to test an arbitrary fitted family rather than only normality. It is less powerful than Anderson--Darling in the tails but gives a straightforward visual diagnostic via the CDF overlay.

    \item \textbf{log\_transform:} Applies $y_t = \log(x_t)$ to stabilize variance in series whose standard deviation grows with the mean, a signature of log-normal or multiplicative processes. Log-transforming a series with exponential growth or multiplicative noise often produces an approximately Gaussian, stationary series amenable to ARIMA. It is also a prerequisite for power-law tail analysis, where log--log linearity of the CCDF is the diagnostic signature.

    \item \textbf{jarque\_bera:} A moment-based normality test with statistic $JB = \frac{n}{6}\left(\gamma_1^2 + \frac{\gamma_2^2}{4}\right)$, which is asymptotically $\chi^2(2)$ under normality. Because it is constructed directly from skewness and kurtosis, it is more powerful than the KS test against symmetric heavy-tailed alternatives (large $\gamma_2$) and skewed alternatives (large $|\gamma_1|$). In practice \method{} uses it as a fast scalar complement to the full \textbf{distribution\_shape\_summary}.

    \item \textbf{ecdf:} Plots the empirical cumulative distribution function $F_n(x) = \frac{1}{n}\sum_{i=1}^n \mathbf{1}[x_i \leq x]$. The ECDF is a non-parametric, unsmoothed estimate of the true CDF and serves as the ground truth against which parametric fits are visually compared. It is the natural complement to the KS test: the test statistic is the maximum vertical gap between the ECDF and the hypothesized CDF curve.

\end{enumerate}

\section{Runtime of \method{}}

\begin{table}[ht]
\centering
\caption{Runtimes by Domain, Dataset, Mode, and Model}
\label{tab:benchmark_results}
\resizebox{\textwidth}{!}{%
\begin{tabular}{llllrrrr}
\toprule
\textbf{Domain} & \textbf{Dataset} & \textbf{Mode} & \textbf{Model} & \textbf{n} & \textbf{Avg Time (s)} & \textbf{Avg Steps} & \textbf{Total Time (s)} \\
\midrule
\multirow{27}{*}{df}
 & \multirow{9}{*}{imf}
   & \multirow{3}{*}{genonly}  & claude\_sonnet46 & 50 & 84.8   & 4.1 & 4238.9   \\
 &  &                           & gpt54\_mini      & 50 & 163.5  & 5.1 & 8173.1   \\
 &  &                           & kimi25           & 50 & 1122.6 & 5.0 & 56131.8  \\
 &  & \multirow{3}{*}{none}     & claude\_sonnet46 & 50 & 64.9   & 4.4 & 3242.6   \\
 &  &                           & gpt54\_mini      & 50 & 31.6   & 5.4 & 1577.9   \\
 &  &                           & kimi25           & 50 & 1073.9 & 5.3 & 53694.8  \\
 &  & \multirow{3}{*}{static}   & claude\_sonnet46 & 50 & 135.2  & 5.8 & 6762.3   \\
 &  &                           & gpt54\_mini      & 50 & 40.1   & 5.8 & 2004.7   \\
 &  &                           & kimi25           & 50 & 1172.4 & 5.5 & 58622.3  \\
\cmidrule{2-8}
 & \multirow{9}{*}{mixed}
   & \multirow{3}{*}{genonly}  & claude\_sonnet46 & 100 & 125.0  & 5.0 & 12503.9  \\
 &  &                           & gpt54\_mini      & 100 & 91.6   & 5.0 & 9163.7   \\
 &  &                           & kimi25           & 100 & 1272.7 & 4.5 & 127267.2 \\
 &  & \multirow{3}{*}{none}     & claude\_sonnet46 & 100 & 103.3  & 5.3 & 10333.3  \\
 &  &                           & gpt54\_mini      & 100 & 21.4   & 5.2 & 2142.7   \\
 &  &                           & kimi25           & 100 & 29.0   & 5.5 & 2905.0   \\
 &  & \multirow{3}{*}{static}   & claude\_sonnet46 & 100 & 152.5  & 5.4 & 15245.5  \\
 &  &                           & gpt54\_mini      & 100 & 24.4   & 5.5 & 2443.5   \\
 &  &                           & kimi25           & 100 & 31.3   & 5.3 & 3127.5   \\
\cmidrule{2-8}
 & \multirow{9}{*}{single}
   & \multirow{3}{*}{genonly}  & claude\_sonnet46 & 50 & 78.3   & 3.7 & 3914.8   \\
 &  &                           & gpt54\_mini      & 50 & 19.4   & 4.6 & 967.9    \\
 &  &                           & kimi25           & 50 & 1166.1 & 3.9 & 58306.0  \\
 &  & \multirow{3}{*}{none}     & claude\_sonnet46 & 50 & 88.6   & 5.2 & 4427.7   \\
 &  &                           & gpt54\_mini      & 50 & 15.6   & 4.2 & 782.2    \\
 &  &                           & kimi25           & 50 & 24.6   & 5.5 & 1232.1   \\
 &  & \multirow{3}{*}{static}   & claude\_sonnet46 & 50 & 101.8  & 4.8 & 5088.8   \\
 &  &                           & gpt54\_mini      & 50 & 26.7   & 4.7 & 1335.1   \\
 &  &                           & kimi25           & 50 & 27.5   & 5.5 & 1375.1   \\
\midrule
\multirow{27}{*}{ts}
 & \multirow{9}{*}{easy}
   & \multirow{3}{*}{genonly}  & claude\_sonnet46 & 50 & 132.3  & 4.8 & 6614.0   \\
 &  &                           & gpt54\_mini      & 50 & 14.8   & 3.3 & 740.8    \\
 &  &                           & kimi25           & 50 & 883.6  & 3.8 & 44181.9  \\
 &  & \multirow{3}{*}{none}     & claude\_sonnet46 & 50 & 51.5   & 3.2 & 2576.6   \\
 &  &                           & gpt54\_mini      & 50 & 11.5   & 3.1 & 576.8    \\
 &  &                           & kimi25           & 50 & 40.1   & 5.9 & 2006.1   \\
 &  & \multirow{3}{*}{static}   & claude\_sonnet46 & 50 & 38.6   & 2.4 & 1927.9   \\
 &  &                           & gpt54\_mini      & 50 & 17.2   & 3.3 & 861.9    \\
 &  &                           & kimi25           & 50 & 50.5   & 5.8 & 2527.5   \\
\cmidrule{2-8}
 & \multirow{9}{*}{gravitational\_chirp}
   & \multirow{3}{*}{genonly}  & claude\_sonnet46 & 50 & 153.0  & 4.5 & 7648.1   \\
 &  &                           & gpt54\_mini      & 50 & 96.7   & 3.3 & 4834.6   \\
 &  &                           & kimi25           & 50 & 986.0  & 4.1 & 49298.0  \\
 &  & \multirow{3}{*}{none}     & claude\_sonnet46 & 50 & 62.6   & 3.5 & 3132.0   \\
 &  &                           & gpt54\_mini      & 50 & 17.0   & 3.4 & 852.0    \\
 &  &                           & kimi25           & 50 & 39.8   & 5.7 & 1989.7   \\
 &  & \multirow{3}{*}{static}   & claude\_sonnet46 & 50 & 87.2   & 3.4 & 4358.0   \\
 &  &                           & gpt54\_mini      & 50 & 22.0   & 3.4 & 1099.4   \\
 &  &                           & kimi25           & 50 & 43.7   & 5.9 & 2184.9   \\
\cmidrule{2-8}
 & \multirow{9}{*}{medium}
   & \multirow{3}{*}{genonly}  & claude\_sonnet46 & 110 & 99.1   & 3.6 & 10902.3  \\
 &  &                           & gpt54\_mini      & 110 & 136.1  & 4.2 & 14968.5  \\
 &  &                           & kimi25           & 110 & 1166.3 & 4.5 & 128297.0 \\
 &  & \multirow{3}{*}{none}     & claude\_sonnet46 & 110 & 53.1   & 3.2 & 5842.3   \\
 &  &                           & gpt54\_mini      & 110 & 12.8   & 2.8 & 1409.4   \\
 &  &                           & kimi25           & 110 & 38.3   & 5.7 & 4208.5   \\
 &  & \multirow{3}{*}{static}   & claude\_sonnet46 & 110 & 49.5   & 2.6 & 5443.2   \\
 &  &                           & gpt54\_mini      & 110 & 15.9   & 2.7 & 1750.4   \\
 &  &                           & kimi25           & 110 & 43.6   & 5.8 & 4791.9   \\
\bottomrule
\end{tabular}%
}
\end{table}

\section{Prompts}
\label{app:vesta_prompts}

This section contains all the prompts used by our method and baselines.

\begin{figure}[h]
\small
\begin{mdframed}[linecolor=gray!50, backgroundcolor=gray!5, linewidth=0.8pt]

\textbf{\method{} Time Series Critique Prompt}

\medskip
You are the VLM critic of \method{}. Inspect the latest fit of the current
Gaussian-process PyMC model and decide whether to refine the kernel
composition further or to terminate the iterative refinement loop. The
prompt covers stationary periodic structure as well as non-stationary
(warped periodic) wave signals.

\medskip
\textbf{Goal:} Determine whether the current GP kernel composition
correctly captures the trend, seasonality, and non-stationarity of the
series, and whether the priors are well-scaled for the observed amplitude
and period.

\medskip
\textbf{Current Model Information}
\begin{itemize}[leftmargin=*, nosep]
    \item Current PyMC code: \texttt{\{current\_pymc\_code\}}
    \item Current kernels: \texttt{\{current\_kernels\}}
    \item Previously tested kernel combinations: \texttt{\{tested\_kernels\}}
    \item History of previously executed tools: \texttt{\{tool\_history\}}
\end{itemize}

\medskip
\textbf{Iteration Summaries:} \texttt{\{summaries\}}

\medskip
\textbf{Important Constraints}
\begin{itemize}[leftmargin=*, nosep]
    \item Evaluate fit quality based on the most recent iteration's tool
          output and visual feedback.
    \item Take the trajectory of past iterations and their metric values
          into account.
    \item You may revisit a previously tested kernel combination, but you
          must use sufficiently different priors.
    \item Do not invoke the same tool more than twice across the run.
    \item Combine kernels by addition only (\texttt{cov = cov1 + cov2}).
    \item \texttt{warped\_periodic} subsumes \texttt{periodic}; do not
          combine them.
\end{itemize}

\medskip
\textbf{Stationary vs.\ Non-Stationary Wave Check:} [\,visual rule for
deciding whether cycle spacing is constant (stationary) or sweeping
(non-stationary), and what kernel that implies\,]

\medskip
\textbf{Kernel Family Guide:} [\,valid kernel names, when each kernel
applies, and how kernels may be combined are listed here\,]

\medskip
\textbf{Prior Specification Rules:} [\,recommended prior families and
scales for length-scale, amplitude, period, warp coefficients, and
observation noise are listed here, with anti-patterns for misuse of
\texttt{pm.Deterministic} and undefined initializer variables\,]

\medskip
\textbf{Proposal Diversity Rules:} [\,the primary proposal must be the best
guess; the remaining proposals must explore structurally different kernel
combinations, including at least one with and one without
\texttt{warped\_periodic} when non-stationary periodicity is suspected\,]

\medskip
\textbf{Diagnostic Results:} \texttt{\{tool\_outputs\}}

\medskip
\textbf{Task:} If the current fit is satisfactory, terminate the loop early
by including the literal token \texttt{COMPLETE} in your description.
Otherwise, propose exactly \texttt{\{num\_proposals\}} diverse revised
models following the rules above. State explicitly whether the
stationary-vs-non-stationary wave check passes and, if non-stationary,
which direction the frequency sweeps.

\medskip
\textbf{Output Format:} Return \textbf{only} a valid JSON object with
\texttt{description} and \texttt{proposals}. Each proposal must include
\texttt{kernels} and \texttt{priors}. Never output anything outside the
JSON object.

\end{mdframed}
\caption{Critique-stage prompt used by \method{} for time series modeling.}
\label{fig:ts-proposal-prompt}
\end{figure}

\begin{figure}[h]
\small
\begin{mdframed}[linecolor=gray!50, backgroundcolor=gray!5, linewidth=0.8pt]

\textbf{\method{} Distribution Fitting Critique Prompt}

\medskip
You are the VLM critic of \method{}. Inspect the latest fit of the current
PyMC model and decide whether to refine it further or to terminate the
iterative refinement loop.

\medskip
\textbf{Goal:} Determine whether the current distribution model correctly
captures the shape, location, scale, and tail behavior of the data.

\medskip
\textbf{Current Model Information}
\begin{itemize}[leftmargin=*, nosep]
    \item Current PyMC code: \texttt{\{current\_pymc\_code\}}
    \item Current distribution family: \texttt{\{current\_family\}}
    \item Previously tested distribution families: \texttt{\{tested\_families\}}
    \item History of previously executed tools: \texttt{\{tool\_history\}}
\end{itemize}

\medskip
\textbf{Iteration Summaries:} \texttt{\{summaries\}}

\medskip
\textbf{Important Constraints}
\begin{itemize}[leftmargin=*, nosep]
    \item Evaluate fit quality based on the most recent iteration's tool
          output and visual feedback.
    \item Take the trajectory of past iterations and their metric values
          into account.
    \item You may revisit a previously tested family, but you must use
          sufficiently different priors.
    \item Do not invoke the same tool more than twice across the run.
    \item Prefer interpretable distributions and avoid unnecessary
          complexity.
\end{itemize}

\medskip
\textbf{Model Fit Diagnostics:} [\,shape capture, peak alignment, tail
behavior, mixture detection, and overfitting checks are listed here\,]

\medskip
\textbf{Distribution Family Guide:} [\,valid single families, mixture
syntax, parametrization reference for each family, support-validation
rules, and family-specific prior constraints are listed here\,]

\medskip
\textbf{Prior Specification Rules:} [\,instructions for reading the
histogram to extract peak/spread/support/tails, translating those
observations into hyperparameter values, and the variable naming convention
for single and mixture priors are listed here\,]

\medskip
\textbf{Proposal Diversity Rules:} [\,the primary proposal must be the best
guess; the remaining proposals must be structurally different alternatives
covering both single distributions and mixtures\,]

\medskip
\textbf{Diagnostic Results:} \texttt{\{tool\_outputs\}}

\medskip
\textbf{Task:} If the current fit is satisfactory, terminate the loop early
by including the literal token \texttt{COMPLETE} in your description.
Otherwise, propose exactly \texttt{\{num\_proposals\}} diverse revised
models following the rules above.

\medskip
\textbf{Output Format:} Return \textbf{only} a valid JSON object with
\texttt{description} and \texttt{proposals}. Each proposal must include
\texttt{distribution\_family}, \texttt{is\_mixture}, and \texttt{priors}.
Never output anything outside the JSON object.

\end{mdframed}
\caption{Critique-stage prompt used by \method{} for distribution fitting.}
\label{fig:dist-proposal-prompt}
\end{figure}

\begin{figure}[h]
\small
\begin{mdframed}[linecolor=gray!50, backgroundcolor=gray!5, linewidth=0.8pt]

\textbf{\method{} Tool Selection Prompt}

\medskip
You are evaluating the current PyMC model fit. The latest fit is shown in
the attached plot.

\medskip
\textbf{Tool Registry:} \texttt{\{tool\_descriptions\}}

\medskip
\textbf{Decision Rules}
\begin{itemize}[leftmargin=*, nosep]
    \item Select exactly one tool from the registry that best diagnoses the
          current fit.
    \item Prefer tools that probe the suspected weakness of the current
          model: \texttt{\{model\_weakness\_examples\}}.
    \item If no existing tool in the registry is adequate, return the
          \texttt{generate\_new\_tool} action and describe what the new tool
          should compute; the tool-creation step will then code it.
    \item You may not invoke the same tool more than twice across the run.
\end{itemize}

\medskip
\textbf{Output Format:} Return \textbf{only} a valid JSON object naming the
selected tool and any arguments it requires.

\medskip
\textbf{Domain-specific values for the placeholders above:}
\begin{itemize}[leftmargin=*, nosep]
    \item \textbf{Distribution fitting} -
          \texttt{\{model\_weakness\_examples\}} = \texttt{peak misalignment, tail mismatch, multimodality, support violation}.
    \item \textbf{Time series modeling} -
          \texttt{\{model\_weakness\_examples\}} = \texttt{missed period, residual structure, amplitude envelope, frequency sweep, anomalies}.
\end{itemize}

\end{mdframed}
\caption{Tool-selection prompt used by the Generate-Tools stage of \method{}.}
\label{fig:vesta-tool-selection-prompt}
\end{figure}

\begin{figure}[h]
\small
\begin{mdframed}[linecolor=gray!50, backgroundcolor=gray!5, linewidth=0.8pt]

\textbf{\method{} Tool Creation Prompt}

\medskip
You are creating a new diagnostic tool that will be added to the tool
registry of \method{} and may be reused in later iterations of this run.
The tool is a Python function executed in a sandbox with access to:
\texttt{numpy as np}, \texttt{pandas as pd}, \texttt{scipy.stats},
\texttt{matplotlib.pyplot as plt}, \texttt{sklearn}\texttt{\{extra\_imports\}}.

\medskip
\textbf{Function Signature}
\begin{itemize}[leftmargin=*, nosep]
    \item Define a single function:
          \texttt{def tool(data: \{data\_handle\}, map\_estimate: dict, family\_or\_kernel: str) -> str}
    \item The return value is the path to a saved figure (PNG) or a short
          textual summary that the VLM critic can read.
\end{itemize}

\medskip
\textbf{Implementation Rules}
\begin{itemize}[leftmargin=*, nosep]
    \item Begin directly with the \texttt{def tool(...)} block. Do
          \textbf{not} include import statements; required modules are
          already in scope.
    \item Use \texttt{map\_estimate} to construct a comparison between the
          observed data and the current model.
    \item Save any figure with \texttt{plt.savefig(...)} and return its
          path; do not call \texttt{plt.show()}.
    \item The tool must run end-to-end without user interaction.
    \item \texttt{\{domain\_specific\_code\_rule\}}
\end{itemize}

\medskip
\textbf{Output Format:} Return \textbf{only} a JSON-parseable dict with key
\texttt{"code"} containing the function as a single-line string with
\texttt{\textbackslash n} for newlines.

\medskip
\textbf{Domain-specific values for the placeholders above:}
\begin{itemize}[leftmargin=*, nosep]
    \item \textbf{Distribution fitting} -
          \texttt{\{extra\_imports\}} = (none beyond the default sandbox);
          \texttt{\{data\_handle\}} = \texttt{np.ndarray};
          \texttt{\{domain\_specific\_code\_rule\}} = ``Visualize the
          empirical and model-implied densities or transformed-data
          summaries; do \textbf{not} fit a new PyMC model inside the tool.''
    \item \textbf{Time series modeling} -
          \texttt{\{extra\_imports\}} = \texttt{, statsmodels.api as sm,
          scipy.signal};
          \texttt{\{data\_handle\}} = \texttt{pd.Series with a DatetimeIndex};
          \texttt{\{domain\_specific\_code\_rule\}} = ``Use the time index
          for any plot's x-axis and align predictions with
          \texttt{data.index}; do \textbf{not} re-fit the GP inside the
          tool.''
\end{itemize}

\end{mdframed}
\caption{Tool-creation prompt used by the Generate-Tools stage of \method{}.}
\label{fig:vesta-tool-creation-prompt}
\end{figure}

\begin{figure}[h]
\small
\begin{mdframed}[linecolor=gray!50, backgroundcolor=gray!5, linewidth=0.8pt]

\textbf{\method{} Summarization Prompt}

\medskip
You are summarizing the latest iteration of a \texttt{\{run\_kind\}} run so
that the next iteration of \method{} can reason over the full refinement
trajectory without consuming the raw intermediate outputs.

\medskip
\textbf{Inputs Available}
\begin{itemize}[leftmargin=*, nosep]
    \item The visual description produced for the current iteration.
    \item The best PyMC model code and its fitted parameters
          (\texttt{map\_estimate}).
    \item The metric value and any auxiliary fit statistics.
    \item The output (image or text) produced by the selected or newly
          created tool.
\end{itemize}

\medskip
\textbf{What the Summary Must Contain}
\begin{itemize}[leftmargin=*, nosep]
    \item A short textual description of what the data look like and what
          structure the current best model captures.
    \item The chosen \texttt{\{model\_field\_name\}} and its key fitted
          parameters as concise key--value pairs.
    \item The metric value for the current best model.
    \item One or two sentences describing what the tool output revealed,
          focusing on remaining mismatch between the model and the data.
    \item \texttt{\{description\_specialization\}}
\end{itemize}

\medskip
\textbf{Style Rules}
\begin{itemize}[leftmargin=*, nosep]
    \item Be concrete: name parameters and numeric values rather than
          describing them abstractly.
    \item Do \textbf{not} re-derive the model or restate prompt
          instructions.
    \item Keep the summary self-contained; it replaces the raw
          per-iteration outputs in subsequent iterations.
\end{itemize}

\medskip
\textbf{Output Format:} Return a JSON object with fields
\texttt{description}, \texttt{\{model\_field\_name\}},
\texttt{map\_estimate}, \texttt{metric}, and \texttt{tool\_output\_summary}.

\medskip
\textbf{Domain-specific values for the placeholders above:}
\begin{itemize}[leftmargin=*, nosep]
    \item \textbf{Distribution fitting} -
          \texttt{\{run\_kind\}} = \texttt{distribution fitting};
          \texttt{\{model\_field\_name\}} = \texttt{distribution\_family};
          \texttt{\{description\_specialization\}} = ``Mention any tail,
          peak, or modality features the current model fails to reproduce.''
    \item \textbf{Time series modeling} -
          \texttt{\{run\_kind\}} = \texttt{time series modeling};
          \texttt{\{model\_field\_name\}} = \texttt{kernels};
          \texttt{\{description\_specialization\}} = ``Mention any trend,
          seasonality, or non-stationarity features the current kernels fail
          to reproduce.''
\end{itemize}

\end{mdframed}
\caption{Prompt used by the Summarize stage of \method{}.}
\label{fig:vesta-summarize-prompt}
\end{figure}


\begin{figure}[h]
\small
\begin{mdframed}[linecolor=gray!50, backgroundcolor=gray!5, linewidth=0.8pt]

\textbf{BoxLM Proposal Prompt for Time Series Modeling}

\medskip
You are a brilliant statistician modeling a time series dataset using
Gaussian processes. Your job is to come up with a GP model that explains
the time series by writing a PyMC probabilistic program. In the first
round you receive only the dataset description; in later rounds you also
receive the previous program along with the critic's hypotheses and
synthesis.

\medskip
\textbf{Inputs}
\begin{itemize}[leftmargin=*, nosep]
    \item Dataset description: \texttt{\{dataset\_description\}}
    \item Column description: \texttt{\{column\_description\}}
    \item Previous round's hypotheses: \texttt{\{prev\_hypotheses\}}
    \item Previous round's synthesis: \texttt{\{prev\_synthesis\}}
\end{itemize}

\medskip
\textbf{Kernel Family Guide}
\begin{itemize}[leftmargin=*, nosep]
    \item \texttt{pm.gp.cov.Linear} - persistent upward or downward drift.
    \item \texttt{pm.gp.cov.Periodic} - smooth repeating sinusoidal
          patterns.
    \item \texttt{pm.gp.cov.ExpQuad} (RBF) - extremely smooth curves.
    \item \texttt{pm.gp.cov.Matern52} - rough or jagged paths.
    \item Combine kernels by addition for independent structures, by
          multiplication for modulation. Do not combine RBF and Matern in
          the same model.
\end{itemize}

\medskip
\textbf{PyMC API Rules}
\begin{itemize}[leftmargin=*, nosep]
    \item Import \texttt{pymc} (not \texttt{pymc3}). Do not use \texttt{sd}
          as an argument; use \texttt{sigma} instead.
    \item Use the variable name \texttt{y\_obs} for the likelihood.
    \item Index columns explicitly from \texttt{observed\_data}; never use
          \texttt{observed\_data.index} to grab raw data.
    \item The model must use Gaussian processes (\texttt{pm.gp}); plain
          regression is not allowed.
    \item Do not use \texttt{pm.Constant} for parameters and do not pass
          \texttt{dims="obs\_id"} for the GP marginal likelihood.
\end{itemize}

\medskip
\textbf{Required Function Skeleton}
\begin{itemize}[leftmargin=*, nosep]
    \item Define a function \texttt{def gen\_model(observed\_data)} that
          loads data via \texttt{observed\_data["column\_name"].values}
          (never a placeholder array literal).
    \item Before the model block, normalize the time axis to
          \texttt{[0, 1]} and center the target values.
    \item Inside \texttt{with pm.Model() as model:} declare the kernel
          priors, build the covariance, define
          \texttt{gp = pm.gp.Marginal(cov\_func=cov)}, and set the
          likelihood
          \texttt{y\_obs = gp.marginal\_likelihood('y\_obs', ..., sigma=sigma)}.
    \item Sample with
          \texttt{trace = pm.sample(1000, tune=500, target\_accept=0.90,
          chains=3, cores=1, random\_seed=rng1,
          idata\_kwargs=\{"log\_likelihood": True\})}.
    \item Then run \texttt{posterior\_predictive =
          pm.sample\_posterior\_predictive(trace, random\_seed=rng2,
          return\_inferencedata=False)}.
    \item Return \texttt{(model, posterior\_predictive, trace)} in that
          order.
\end{itemize}

\medskip
\textbf{Response Structure}
\begin{itemize}[leftmargin=*, nosep]
    \item First, sketch a high-level GP model for the data and explicitly
          cite which previous-round hypotheses (if any) the sketch
          addresses.
    \item Then write the full program in a Python code block, with comments
          on lines that address specific hypotheses.
\end{itemize}

\end{mdframed}
\caption{Proposal prompt used by the BoxLM baseline for time series modeling.}
\label{fig:boxlm-proposal-time-series-prompt}
\end{figure}

\begin{figure}[h]
\small
\begin{mdframed}[linecolor=gray!50, backgroundcolor=gray!5, linewidth=0.8pt]

\textbf{BoxLM Proposal Prompt for Distribution Fitting}

\medskip
You are a brilliant statistician modeling a dataset. Your job is to come up
with a generative model that explains the true data by writing a PyMC
probabilistic program. In the first round you receive only the dataset
description; in later rounds you also receive the previous program along
with the critic's hypotheses and synthesis.

\medskip
\textbf{Inputs}
\begin{itemize}[leftmargin=*, nosep]
    \item Dataset description: \texttt{\{dataset\_description\}}
    \item Column description: \texttt{\{column\_description\}}
    \item Previous round's hypotheses: \texttt{\{prev\_hypotheses\}}
    \item Previous round's synthesis: \texttt{\{prev\_synthesis\}}
\end{itemize}

\medskip
\textbf{PyMC API Rules}
\begin{itemize}[leftmargin=*, nosep]
    \item Import \texttt{pymc} (not \texttt{pymc3}). Do not use \texttt{sd}
          as an argument; use \texttt{sigma} instead.
    \item Use the variable name \texttt{y\_obs} for the likelihood.
    \item Index columns explicitly from \texttt{observed\_data}; never use
          \texttt{observed\_data.index} to grab raw data.
    \item Wrap every input feature in a \texttt{pm.MutableData} container
          with \texttt{dims="obs\_id"}, and pass \texttt{dims="obs\_id"} to
          the likelihood as well.
    \item Do not use \texttt{pm.Constant} for parameters and do not use
          Gaussian processes for this task.
    \item For Student-T, give \texttt{nu} a continuous positive prior such
          as \texttt{pm.Exponential} or \texttt{pm.HalfNormal}.
\end{itemize}

\medskip
\textbf{Required Function Skeleton}
\begin{itemize}[leftmargin=*, nosep]
    \item Define a function \texttt{def gen\_model(observed\_data)} that
          loads data via \texttt{observed\_data["column\_name"].values}
          (never a placeholder array literal).
    \item Inside \texttt{with pm.Model() as model:} declare the
          \texttt{MutableData} containers, the priors, and the likelihood
          \texttt{y\_obs}.
    \item Sample with
          \texttt{trace = pm.sample(1000, tune=500, target\_accept=0.90,
          chains=3, cores=1, random\_seed=rng1,
          idata\_kwargs=\{"log\_likelihood": True\})}.
    \item Then run \texttt{posterior\_predictive =
          pm.sample\_posterior\_predictive(trace, random\_seed=rng2,
          return\_inferencedata=False)}.
    \item Return \texttt{(model, posterior\_predictive, trace)} in that
          order.
\end{itemize}

\medskip
\textbf{Response Structure}
\begin{itemize}[leftmargin=*, nosep]
    \item First, sketch a high-level probabilistic program for the data and
          explicitly cite which previous-round hypotheses (if any) the
          sketch addresses.
    \item Then write the full program in a Python code block, with comments
          on lines that address specific hypotheses.
\end{itemize}

\end{mdframed}
\caption{Proposal prompt used by the BoxLM baseline for distribution fitting.}
\label{fig:boxlm-proposal-distribution-fitting-prompt}
\end{figure}

\begin{figure}[h]
\small
\begin{mdframed}[linecolor=gray!50, backgroundcolor=gray!5, linewidth=0.8pt]

\textbf{BoxLM Critic Prompt}

\medskip
You are a brilliant statistician specializing in critiquing and proposing
revisions of probabilistic models. Your colleague has produced a list of
PyMC programs that hypothesize \texttt{\{model\_class\}} for the data; each
program has been fit, plotted against the actual data, and scored with
ELPD-LOO (higher is better). Your job is to figure out whether the
proposed programs are consistent with the actual data by comparing the
posterior predictive plots against the actual data, and to provide
modeling-related guidance for the next round of revision.

\medskip
\textbf{Inputs}
\begin{itemize}[leftmargin=*, nosep]
    \item Dataset description: \texttt{\{dataset\_description\}}
    \item Column description: \texttt{\{column\_description\}}
    \item Expert context: \texttt{\{expert\_context\}}
    \item Programs from this round, each with its ELPD-LOO score and a
          posterior-predictive plot overlaid on the actual data.
    \item Hypotheses and synthesis from previous rounds (if any), which
          may be revised, removed, or extended.
\end{itemize}

\medskip
\textbf{Instructions}
\begin{itemize}[leftmargin=*, nosep]
    \item Look at every posterior-predictive plot and explicitly describe
          the visual properties of the model versus the data.
    \item State natural-language hypotheses for any discrepancies between
          the proposed programs and the actual data, prioritizing the
          larger discrepancies.
    \item Using the programs and their ELPD-LOO scores, write a synthesis
          of which strategies and modeling choices did and did not work.
          The synthesis is passed back to your colleague, so make it as
          informative as possible. Restrict suggestions to modeling; do
          not suggest collecting more data.
    \item If hypotheses or a synthesis from previous rounds are provided,
          revise them based on the new evidence and briefly explain any
          additions or removals.
\end{itemize}

\medskip
\textbf{Output Format}\\
Respond with two fenced blocks in this exact order:

\begin{verbatim}
``` Hypotheses
    * Hypothesis
    * Hypothesis
```

```Synthesis
    (Your synthesis here)
```
\end{verbatim}

\medskip
\textbf{Domain-specific values}
\begin{itemize}[leftmargin=*, nosep]
    \item For distribution fitting, \texttt{\{model\_class\}} is
          \emph{generative models}.
    \item For time series modeling, \texttt{\{model\_class\}} is
          \emph{Gaussian process models}.
\end{itemize}

\end{mdframed}
\caption{Critic prompt used by the BoxLM baseline for both distribution fitting and time series modeling.}
\label{fig:boxlm-critic-prompt}
\end{figure}


\begin{figure}[h]
\small
\begin{mdframed}[linecolor=gray!50, backgroundcolor=gray!5, linewidth=0.8pt]

\textbf{PyVision Agent Prompt}

\medskip
You are an agent; please keep going until the user's query is completely
resolved before ending your turn and yielding back to the user. Only
terminate your turn when you are sure that the problem is solved. Solve
the following problem step by step. You now have the ability to
selectively write executable Python code to enhance your reasoning
process. The Python code will be executed by an external sandbox.

\medskip
You MUST plan extensively before each function call, and reflect
extensively on the outcomes of the previous function calls. Do not do
this entire process by making function calls only, as this can impair
your ability to solve the problem and think insightfully.

\medskip
For all the provided images, in order, the \emph{i}-th image has already
been read into the global variable \texttt{image\_clue\_i} using
\texttt{PIL.Image.open()}. When writing Python code, you can directly use
these variables without needing to read them again.

\medskip
Since you are dealing with the vision-related task, you MUST use the
Python tool (e.g., the \texttt{matplotlib} library) to analyze or
transform the image whenever it could improve your understanding or aid
your reasoning. This includes but is not limited to zooming in, rotating,
adjusting contrast, computing statistics, or isolating features. When you
use \texttt{matplotlib} to visualize data or further process images, use
\texttt{plt.show()} to display them; there is no need to save them. Do
not use image-processing libraries like \texttt{cv2} or \texttt{PIL}. If
you want to check the value of a variable, you MUST use \texttt{print()}
to check it.

\medskip
\textbf{Code-execution Format}\\
The output (wrapped in
\texttt{<interpreter>output\_str</interpreter>}) is returned to aid your
reasoning. Each Python snippet must be a complete script including
necessary imports, wrapped exactly as:

\begin{verbatim}
<code>
```python
code snippet
```
</code>
\end{verbatim}

\medskip
\textbf{Image Information}\\
\texttt{Image Width: \{width\}; Image Height: \{height\}}

\medskip
\textbf{User Query}\\
\texttt{\{query\}}

\medskip
\textbf{Final Answer Format}\\
The last part of your response must be in the following format:

\begin{verbatim}
<answer>
\boxed{'The final answer goes here.'}
</answer>
\end{verbatim}

\medskip
\textbf{Domain-specific values}
\begin{itemize}[leftmargin=*, nosep]
    \item For distribution fitting, \texttt{\{query\}} instructs the agent
          to propose and fit a PyMC probabilistic program for the
          empirical distribution shown in the histogram image, using the
          same PyMC coding guidelines as the \method{} agent.
    \item For time series modeling, \texttt{\{query\}} instructs the agent
          to propose and fit a PyMC Gaussian process program for the
          series shown in the time series plot, using the same PyMC
          coding guidelines as the \method{} agent.
\end{itemize}

\end{mdframed}
\caption{Agent prompt used by the PyVision baseline for both distribution fitting and time series modeling.}
\label{fig:pyvision-agent-prompt}
\end{figure}


\begin{figure}[h]
\small
\begin{mdframed}[linecolor=gray!50, backgroundcolor=gray!5, linewidth=0.8pt]

\textbf{Coding Agent Prompt for Time Series Modeling}

\medskip
You are solving a time-series problem. Infer the structural components
that most likely generated the signal. The initial data plot is
\texttt{initial\_plot.png}.

\medskip
\textbf{Environment}
\begin{itemize}[leftmargin=*, nosep]
    \item The dataset is \texttt{data.pkl}, a pickled pandas Series
          indexed by time. Read it directly with
          \texttt{pickle.load(open('data.pkl', 'rb'))}.
    \item You have access to Bash, Read, Write, Edit, Glob, and Grep.
          PyMC (\texttt{pm}), NumPy, pandas, SciPy, and matplotlib are
          installed.
    \item Run Python with \texttt{\{python\_path\}} (the interpreter that
          has these packages).
    \item Everything you need is in this directory. All file access is
          confined here; attempts to read or write outside this directory
          will be blocked, so do not attempt to access the broader
          filesystem.
\end{itemize}

\medskip
Use the \texttt{fit\_pymc} MCP tool to fit and evaluate PyMC models
against this data. When you call \texttt{fit\_pymc}, you may pass a
\texttt{distribution\_family} argument (a list of the structural
components you used) to label your proposal.

\medskip
\textbf{Code Contract for \texttt{fit\_pymc}}
\begin{itemize}[leftmargin=*, nosep]
    \item \texttt{data} is a pandas Series indexed by time, already in
          scope. Do not redefine it.
    \item Define a PyMC model in \texttt{with pm.Model() as model:}.
    \item Define \texttt{map\_estimate = pm.find\_MAP()}.
    \item Define \texttt{trend} as the predicted mean series over the
          observed times.
\end{itemize}

\end{mdframed}
\caption{Coding-agent baseline prompts for time series.}
\label{fig:coding-agent-time-series-prompt}
\end{figure}

\begin{figure}[h]
\small
\begin{mdframed}[linecolor=gray!50, backgroundcolor=gray!5, linewidth=0.8pt]

\textbf{Coding Agent Prompt for Distribution Fitting}

\medskip
You are solving a distribution-fitting problem. Infer the distribution
family (or mixture) that most likely generated the data. The initial data
plot is \texttt{initial\_plot.png}.

\medskip
\textbf{Environment}
\begin{itemize}[leftmargin=*, nosep]
    \item The dataset is \texttt{data.pkl}, a pickled numpy array. Read it
          directly with \texttt{pickle.load(open('data.pkl', 'rb'))}.
    \item You have access to Bash, Read, Write, Edit, Glob, and Grep.
          PyMC (\texttt{pm}), NumPy, SciPy, and matplotlib are installed.
    \item Run Python with \texttt{\{python\_path\}} (the interpreter that
          has these packages).
    \item Everything you need is in this directory. All file access is
          confined here; attempts to read or write outside this directory
          will be blocked, so do not attempt to access the broader
          filesystem.
\end{itemize}

\medskip
Use the \texttt{fit\_pymc} MCP tool to fit and evaluate PyMC models
against this data. When you call \texttt{fit\_pymc}, you may pass a
\texttt{distribution\_family} argument (a list of distribution family
names) to label your proposal.

\medskip
\textbf{Code Contract for \texttt{fit\_pymc}}
\begin{itemize}[leftmargin=*, nosep]
    \item \texttt{data} is a 1-D numpy array, already in scope. Do not
          redefine it.
    \item Define a PyMC model in \texttt{with pm.Model() as model:}.
    \item Define \texttt{map\_estimate = pm.find\_MAP()}.
\end{itemize}

\end{mdframed}
\caption{Coding-agent baseline prompts for distribution fitting.}
\label{fig:coding-agent-distribution-fitting-prompt}
\end{figure}

\section{Limitations}
While \method{} demonstrates strong performance across \data{}, several limitations remain. First, \method{} relies on VLMs to interpret multi-panel diagnostic visualizations, and current VLMs struggle to fully unpack and reason over complex visual outputs. This bottleneck directly limits the gains from dynamic tool creation, as the most sophisticated generated tools contain multi-panel images that compose multiple functions. This gap likely accounts for the remaining 
performance differential between \method{}-Dynamic and \method{}-Expert.

Second, the iterative nature of \method{} incurs non-trivial computational cost, as each 
iteration requires fitting up to $p$ PyMC models via MAP estimation. This limits 
scalability to very large datasets.

Finally, while \data{} covers two domains central to scientific modeling and includes 
real-world \protect\Astro{} tasks, the benchmark is synthetically generated. Performance 
on fully observational scientific datasets, where ground-truth parameters are unknown, 
remains an important avenue for future evaluation.

\newpage

\newpage

\end{document}